\pdfoutput=1

\documentclass[11pt]{article}

\usepackage[]{acl}

\usepackage[T1]{fontenc}

\usepackage[utf8]{inputenc}

\usepackage{microtype}
\usepackage{times}
\usepackage{latexsym}
\usepackage{amssymb}
\usepackage{amsfonts}
\usepackage{amsmath} 
\usepackage{amsthm} 
\usepackage{booktabs}
\usepackage{enumerate}
\usepackage{graphicx}
\usepackage{subfigure}
\usepackage{xspace}
\usepackage{float}
\usepackage{bbm}
\usepackage{bm}
\usepackage{multirow}
\usepackage{booktabs}
\usepackage{color}
\usepackage{framed}
\usepackage{stfloats}
\usepackage{iitem}
\usepackage{makecell}
\usepackage{mathrsfs}
\usepackage[normalem]{ulem}
\useunder{\uline}{\ul}{}
\usepackage{url}
\newcommand{\paratitle}[1]{\vspace{1.5ex}\noindent\textbf{#1}}
\newcommand{\ie}{\emph{i.e.,}\xspace}

\newcommand{\eg}{\emph{e.g.,}\xspace}

\newcommand{\ignore}[1]{}
\newcommand{\tabincell}[2]{\begin{tabular}{@{}#1@{}}#2\end{tabular}}

\usepackage{xcolor}
\definecolor{tOrange}{RGB}{255,165,0}
\definecolor{tBlue}{RGB}{24,116,205}
\definecolor{tPink}{RGB}{255,20,147}
\definecolor{tGreen}{RGB}{50,205,50}
\definecolor{tGold}{RGB}{255,215,0}

\title{ElitePLM: An Empirical Study on General Language Ability Evaluation of Pretrained Language Models}

\author{
	Junyi Li\textsuperscript{\rm{1,5}}, 
	Tianyi Tang\textsuperscript{\rm{1}}, 
	Zheng Gong\textsuperscript{\rm{1}}, 
	Lixin Yang\textsuperscript{\rm{2}},
	Zhuohao Yu\textsuperscript{\rm{2}},
	Zhipeng Chen\textsuperscript{\rm{2}},\\
	\textbf{Jingyuan Wang}\textsuperscript{\rm{3,4}},
     \textbf{Wayne Xin Zhao}\textsuperscript{\rm{1,5}\thanks{\ \ Corresponding author}\ } {\rm and}
	\textbf{Ji-Rong Wen}\textsuperscript{\rm{1,5}} \\
	\textsuperscript{1}Gaoling School of Artificial Intelligence, Renmin University of China \\
	\textsuperscript{2}Renmin University of China \quad \textsuperscript{3}Peng Cheng Laboratory \\
	\textsuperscript{4}School of Computer Science and Engineering, Beihang University \\
	\textsuperscript{5}Beijing Key Laboratory of Big Data Management and Analysis Methods \\
	\texttt{\{lijunyi,steven\_tang\}@ruc.edu.cn} \quad \texttt{batmanfly@gmail.com} \\ 
}

\begin{document}
\maketitle

\begin{abstract}
Nowadays, pretrained language models (PLMs) have dominated the majority of NLP tasks. While, little research has been conducted on systematically evaluating the language abilities of PLMs. In this paper, we present a large-scale empirical study on gen\underline{E}ral \underline{l}anguage ab\underline{i}li\underline{t}y \underline{e}valuation of \underline{PLM}s (\textbf{ElitePLM}). In our study, we design four evaluation dimensions, \ie memory, comprehension, reasoning, and composition, to measure ten widely-used PLMs within five categories. Our empirical results demonstrate that: (1) PLMs with varying training objectives and strategies are good at different ability tests; (2) fine-tuning PLMs in downstream tasks is usually sensitive to the data size and distribution; (3) PLMs have excellent transferability between similar tasks. Moreover, the prediction results of PLMs in our experiments are released as an open resource for more deep and detailed analysis on the language abilities of PLMs. This paper can guide the future work to select, apply, and design  PLMs for specific tasks. We have made all the details of experiments publicly available at~\url{https://github.com/RUCAIBox/ElitePLM}.

\end{abstract}

\section{Introduction}
\label{sec-intro}

Recent years have featured a trend towards Transformer~\citep{VaswaniSPUJGKP17} based pretrained language models (PLMs) in natural language processing (NLP) systems. By being pretrained on massive unlabeled text, PLMs can be directly fine-tuned on downstream tasks, entirely removing the need for  task-specific architectures~\citep{radford2018improving}. This paradigm has led to significant progress on many challenging NLP tasks such as reading comprehension~\citep{DevlinCLT19} and text generation~\citep{brown2020language}.

With rising new state-of-the-art results that approach or surpass human performance on several tasks, it is a non-trivial research  topic about how to systematically evaluate the language abilities of PLMs from a wide range of perspectives.
Given a wide range of publicly released PLMs, it is particularly useful to derive principles or guidelines for selecting suitable PLMs for specific downstream tasks.
However, existing works either target some single ability~\citep{TalmorEGB20,ZhouZCH20}, or consider a simple mixture of multiple (small-scale) tasks that lack a comprehensive design and test~\citep{WangSMHLB19,CLUEbenchmark}. 
There has been no detailed and systematic analysis of PLM's abilities in large-scale NLP tasks. To fill the gap of PLMs evaluation, we introduce the gen\textbf{\underline{E}}ral \textbf{\underline{l}}anguage ab\textbf{\underline{i}}li\textbf{\underline{t}}y \textbf{\underline{e}}valuation (\textbf{ElitePLM}) for empirically and systematically assessing the general language abilities of PLMs. 

The ideal goal behind PLMs is to create a human-like machine learner where it can understand the language and then perform any specific task related to language. In cognitive science, Wechsler Adult Intelligence Scale (WAIS)~\citep{kaufman2005assessing} is the most commonly used intelligence quotient (IQ) test for measuring the intelligence and cognitive ability of humans. This test would assess the level of individuals on verbal comprehension, perceptual reasoning, working memory, and processing speed. Thus, by imitating the intelligence test on humans, we design four evaluation dimensions in ElitePLM for measuring the abilities of PLMs, including \textit{memory}, \textit{comprehension}, \textit{reasoning}, and \textit{composition}. Following previous works~\citep{ZhouZCH20,WangSMHLB19}, for each ability in ElitePLM, we elaborate and select multiple representative tasks (\eg question answering for the comprehension ability) and commonly-used benchmarks (\eg GLUE and SQuAD) to quantitatively evaluate the performance of PLMs. These results can serve as numerical explanations of PLMs at a specific ability.

In human intelligence tests, the background of participants (\eg gender, race, and occupation) should be as much as diverse. Thus, in ElitePLM, we select a diversity of PLMs to conduct generalized and meaningful comparisons. According to training objectives, PLMs can be divided into three types: bidirectional LMs (\eg BERT~\citep{DevlinCLT19}) for natural language understanding (NLU), unidirectional LMs (\eg GPT~\citep{radford2019language}) for natural language generation (NLG), and hybrid LMs (\eg UniLM~\citep{00040WWLWGZH19}) for combining these two paradigms. Furthermore, knowledge-enhanced LMs (\eg ERNIE~\citep{ZhangHLJSL19}) and text-to-text LMs (\eg T5~\citep{RaffelSRLNMZLL20}) also emerge as important branches of PLMs. 
Considering the variety, we finally select ten widely-used PLMs within the above five categories and evaluate their abilities on four dimensions. We show the comparisons of these PLMs in Table~\ref{tab:models} of Appendix~\ref{configuration}.


From the ability test results, we have three salient findings. First, PLMs with varying pretraining objectives and strategies are good at different kinds of downstream tasks. Specifically, we observe that bidirectional LMs like BERT and pretraining strategies like larger training batches in RoBERTa are helpful for memorizing pretraining corpora; permutation language modeling in XLNet is beneficial for modeling the bidirectional context in language comprehension; inter-sentence coherence objective in ALBERT is suitable for sentence-level reasoning tasks; text-to-text LMs using denoising objective like BART perform better in short text generation. Second, when fine-tuning PLMs in downstream tasks, their performance is typically sensitive to the data distribution in fine-tuning stage, which can be addressed by incorporating intermediate datasets or tasks to alleviate such a discrepancy. Third, PLMs have excellent transferability between similar tasks, especially reasoning tasks. This finding will inspire future researchers to leverage data-rich tasks for improving data-scarce tasks. 
For more clarity, we illustrate the impact level of each factor for PLMs' abilities in Table~\ref{tab:score} of Appendix~\ref{configuration}.

Besides ElitePLM being an evaluation benchmark of PLMs' language ability, more importantly, the predicted results of ElitePLM can be used as an open resource for more depth and granularity in analyzing PLMs performance on each ability. For example, we further analyze the comprehension test results of PLMs across answer types in QA tasks. The analysis shows that PLMs are good at simple single-token answers such as dates but more challenged on intricate phrase answers. Moreover, by analyzing human test and Turing test results on composition, we observe that summaries with high accuracy are more likely to pass the Turing test while rich information is more important for story generation. Overall, ElitePLM can act as an analysis tool to gain more insight into PLMs. We show the details of our used datasets and predicted outputs of PLMs in Appendix~\ref{statistics}.

This paper is intended to help establish sound principles for choosing, applying, interpreting and designing PLMs for NLP tasks in practical settings. We have released the code and predicted results of each ability experiment, providing the research and industry community with off-the-shelf tools to evaluate and analyze their PLMs.

\section{ElitePLM}
\label{sec-glae}

In this section, we will detail these four kinds of language abilities, \ie memory, comprehension, reasoning, and composition, in ElitePLM.


\paratitle{Memory Ability.} Memory is the most basic ability of humanity, involved in how much information we recall throughout our lives~\citep{miyake1999models}. By analogy, ElitePLM will measure how much knowledge and language patterns PLMs have memorized in pretraining, as assessed by tests of recalling words based on contexts. Based on the memorized information, PLMs can effectively adapt to downstream tasks for understanding and reasoning about the similar context in a specific text.
On the other hand, efficiency is also a critical aspect of memory ability for PLMs learning from new data distribution in the fine-tuning stage. Thus, besides recalling words, we also compare the memory efficiency of PLMs in terms of memorizing the given new information. 

\paratitle{Comprehension Ability.} Comprehension is an intricate and multifaceted ability. It typically consists of understanding a text’s vocabulary, background knowledge of a specific topic, and comprehension of its linguistic structures like grammar~\citep{cain2008children}. In particular, background (prior) knowledge is used to comprehend a special situation, lesson, or text. 
For example, readers should be aware of the background knowledge of dog behavior when reading a text about dog training.
In ElitePLM, we will assess PLMs' comprehension ability from three aspects, \ie vocabulary, background knowledge, and linguistic structures. 

\paratitle{Reasoning Ability.} Based on the comprehension of a text, reasoning ability refers to the power of the processes and strategies used in drawing inferences, reaching conclusions, arriving at solutions, and making decisions~\citep{kyllonen1990reasoning}. 
In ElitePLM, we mainly focus on three types of reasoning abilities. In detail, commonsense reasoning requires PLMs to draw inferences using commonsense knowledge about the world, like the fact that ``\textit{matches}'' plus ``\textit{logs}'' usually equals ``\textit{fire}''~\citep{SapSBCR20}; Note that subtle differences exist between commonsense knowledge and background knowledge in comprehension ability. Commonsense knowledge is broadly defined as the total accumulation of facts and information that a person has gained from previous experiences. 
Deductive reasoning involves PLMs drawing conclusions from a set of given premises in the form of categorical syllogisms (\eg all $x$ are $y$) or symbolic logic (\eg if $p$ then $q$)~\citep{johnson1999deductive}; Abductive reasoning involves reaching the most likely explanation for a set of facts, such as a scientific theory to explain a set of empirical findings~\citep{walton2014abductive}.

\paratitle{Composition Ability.} 
In the literature~\cite{connors1997composition}, composition is a highly intelligent and synthetic process where a writer assembles words and sentences to create a coherent and meaningful work (\eg poem, music, and novel) from scratch, which closely resembles to the text generation task in NLP~\citep{berninger1999coordinating}. 
Therefore, in ElitePLM, we introduce several text generation tasks to evaluate the composition ability of PLMs, including story generation, text summarization, and question generation. Note that, story generation is a representative composition task which needs PLMs to not only comprehend the given story background, but also reason about and create reasonable and coherent story endings~\citep{LewisDF18}. During the composition process, PLMs should include a good vocabulary, grammar, spelling, and punctuation knowledge, and deliberate the text structure.

\section{Experiments}

In this section, we first set up baselines, and then report the results and analysis on four ability tests.

\subsection{Models}

As mentioned before, we compare the performance of ten publicly released PLMs from five categories: (1) \textit{Bidirectional LMs}: BERT~\citep{DevlinCLT19}, RoBERTa~\citep{liu2019roberta}, and ALBERT~\citep{LanCGGSS20}; (2) \textit{Unidirectional LMs}: GPT-2~\citep{radford2019language}; (3) \textit{Hybrid LMs}: XLNet~\citep{YangDYCSL19} and UniLM~\citep{00040WWLWGZH19}; (4) \textit{Knowledge-enhanced LMs}: ERNIE~\citep{ZhangHLJSL19}; (5) \textit{Text-to-Text LMs}: BART~\citep{LewisLGGMLSZ20}, T5~\citep{RaffelSRLNMZLL20}, and ProphetNet~\citep{QiYGLDCZ020}.
We implement these models and ability tests mostly on huggingface~\citep{wolf-etal-2020-transformers}, fairseq~\citep{ott2019fairseq}, and jiant~\citep{phang2020jiant}. To reflect the true level of language abilities, we adopt the best hyper-parameter values reported in their original papers for each PLM.

\subsection{Memory Tests}
\label{sec-memory}

\paratitle{Datasets and Metrics.} The goal of memory tests is to assess how much knowledge and language patterns PLMs have memorized during pretraining. For this purpose, we adopt two datasets for evaluation, \ie LAMA~\citep{petroni2019language} and English Wikipedia (2,500M words). Specifically, LAMA is a knowledge probe corpus containing a set of knowledge facts, where facts are either subject-relation-object triples or question-answer pairs. Each fact is converted into a cloze statement where the subject or object entity is masked. Wikipedia is one of the widely-used pretraining corpora for our selected PLMs (except GPT-2 and T5). Therefore, to conduct a fair comparison, we continuously train GPT-2 and T5 on Wikipedia using their pretraining objectives. Similar to LAMA, we randomly sample 100,000 texts from Wikipedia and then mask a proportion of $15\%$ tokens following BERT. By querying PLMs with the missing tokens on Wikipedia and LAMA, we can test the language patterns and factual knowledge in PLMs' memory. For metrics, we use \textit{Mean Precision at One ($P$@$1$)} of predicting missing tokens. 
For efficiency, we measure it as the performance \emph{w.r.t.} the number of training epochs: the more efficient a model is, the fewer epochs to achieve a reference performance. 

\begin{figure}[t]
	\centering
	\subfigure[Google-RE]{
		\centering
		\includegraphics[width=0.22\textwidth]{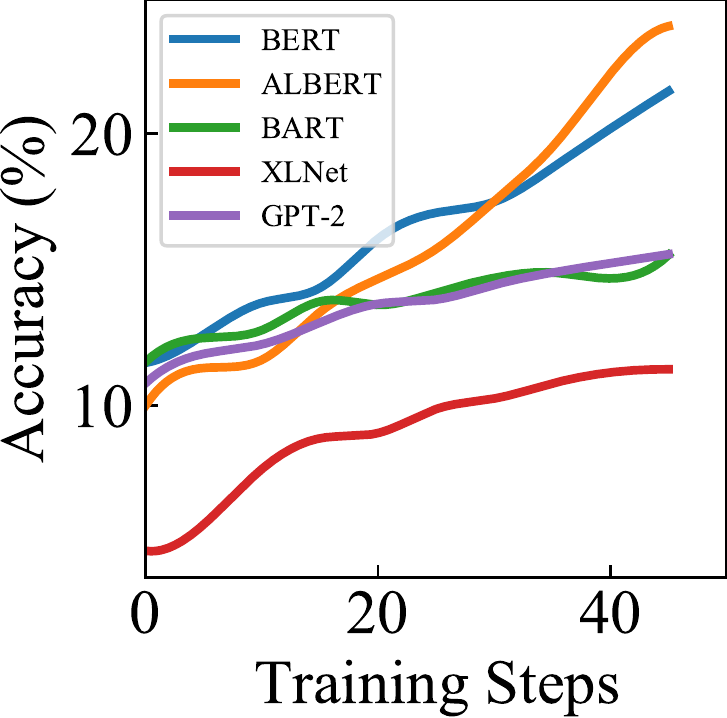}
	}
	\subfigure[T-REx]{
		\centering
		\includegraphics[width=0.22\textwidth]{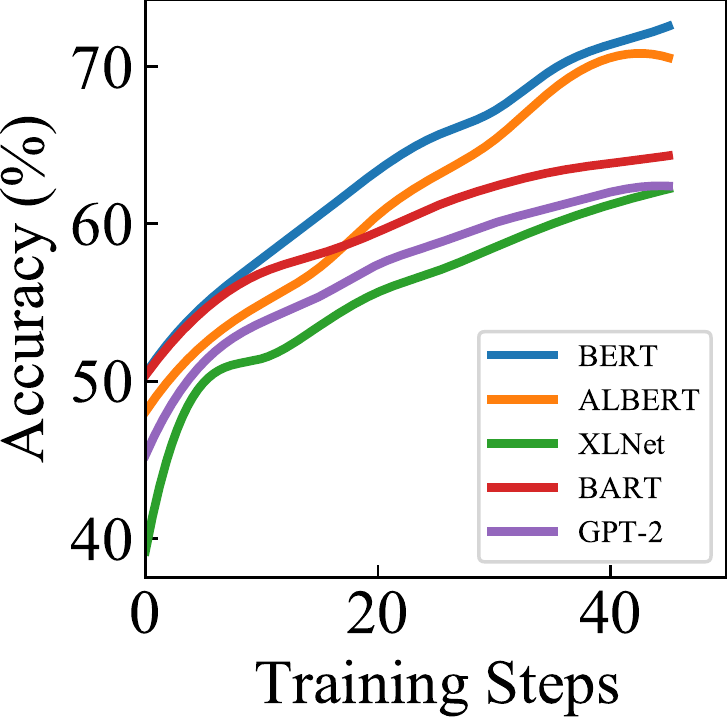}
	}
	\centering
	\caption{Memory efficiency ($P$@1) of five PLMs on Google-RE and T-REx datasets.}
	\label{fig:memory-efficiency}
\end{figure} 

\begin{table*}[t]
 \small
 \centering
 \begin{tabular}{lrrrrrrrrrr}
  \toprule
  		\multirow{2.5}*{\textbf{Models}}& \multicolumn{3}{c}{\textbf{Bidirectional}} & \multicolumn{1}{c}{\textbf{Uni.}}& \multicolumn{2}{c}{\textbf{Hybrid}} & \multicolumn{1}{c}{\textbf{KE}}  & \multicolumn{3}{c}{\textbf{Text-to-Text}}  \\
  		\cmidrule(lr){2-4}\cmidrule(lr){5-5}\cmidrule(lr){6-7}\cmidrule(lr){8-8}\cmidrule(lr){9-11}
         & BERT          & RoBERTa       & ALBERT     & GPT-2 & XLNet  & UniLM    & ERNIE &   T5       & BART  & ProphetNet \\ 
        \midrule
        \textbf{Vocab Size} & 28996         & 50265         & 30000      & 50257 & 32000 & 28996         & 28996 & 32100 & 50295 & 30522 \\
        \cmidrule{1-11}
        \textbf{LAMA} \\
        \makecell[r]{Google-RE}     & \textbf{11.0} & 7.1           & 3.3        & 3.9   & {\ul 10.0}   &  9.6     & 1.3   & 4.0   & 9.4   & 0.1   \\
        \makecell[r]{T-REx}   &  \textbf{29.2} & 23.9          & 21.0       & 12.0  & {\ul 28.9}  &  28.4    & 13.4  & 21.7  & 15.8  & 1.1   \\
        \makecell[r]{ConceptNet}   &  19.1          & \textbf{21.6} & {\ul 20.0} & 6.4   & 19.5   & 18.3          & 13.0   & 17.1  & 7.7   & 0.3   \\
        \makecell[r]{SQuAD}     & 17.0          & \textbf{21.0} & 20.6 & 5.6   & {\ul 20.8}   & 17.4          & 8.1   & 11.7  & 3.1   & 0.7   \\
        \cmidrule{1-11}
        \textbf{Wikipedia}  & 70.9          & {\ul 71.1}    & 63.9       & 42.7  & 68.7  & \textbf{71.5} &   45.7   & 65.0  & 47.8  & 31.3 \\
  \bottomrule
 \end{tabular}
 \caption{Memory test results on LAMA and Wikipedia datasets (test set). These results are based on the \textsc{Large} version of each PLM and more results can be found in the Appendix~\ref{memory}. Bold and underlined fonts denote the best and the second best performance of a PLM (the same as below).}
 \label{tab:memory-main}
\end{table*}

\begin{table*}[t]
 \small
 \centering
 \begin{tabular}{c c r r r r r}
  \toprule
  		\textbf{Relation} & \textbf{Template} & \textbf{BERT} & \textbf{RoBERTa} & \textbf{GPT-2} & \textbf{BART} & \textbf{T5} \\
        \midrule
        \multirow{3}*{<\texttt{[X]}, place\_of\_death, \texttt{[Y]}>} & \texttt{[X]} died in \texttt{[MASK]}. & 13.98 & 0.46 & 0.15 & 11.09 & 4.19 \\
        & \texttt{[X]} passed away in \texttt{[MASK]}. &  13.46 & 0.46 & 0.62 & 3.54 & 1.51 \\
        & \texttt{[X]}'s place of death was \texttt{[MASK]}. & 3.27 & 0.00 & 0.00 & 0.00 & 1.51 \\
        \midrule
        \multirow{3}*{<\texttt{[X]}, place\_of\_birth, \texttt{[Y]}>} & \texttt{[X]} was born in \texttt{[MASK]}. & 16.07 & 12.52 & 7.53 & 14.77 & 6.32 \\
        & \texttt{[X]} was born in the place of \texttt{[MASK]}. & 2.83 & 1.29 & 0.00 & 0.00 & 1.39 \\
        & \texttt{[X]}'s place of birth was \texttt{[MASK]}. & 12.16 & 1.87 & 0.00 & 0.00 & 3.12 \\
  \bottomrule
 \end{tabular}
 \caption{The impact of template on eliciting PLMs' stored knowledge.}
 \label{tab:memory-prompt}
\end{table*}

\paratitle{Results and Analysis}. To evaluate how much text PLMs have recalled in pretraining, we directly test PLMs using Wikipedia and LAMA without fine-tuning, similar to zero-shot learning. The results on $P$@$1$ metric are shown in Table~\ref{tab:memory-main}. Compared with bidirectional and hybrid LMs (\eg BERT and XLNet), GPT-2 uses auto-regressive self-attention where every token can only attend to the context to its left. This unidirectional training objective naturally limits the performance of GPT-2 in terms of memorizing information. It has been previously reported that PLMs can remember more information by scaling up the model size~\cite{brown2020language}. However, in our tests, BART-large (400M) achieves worse results than RoBERTa-base (125M) with the same training corpus and similar vocabulary sizes (50,295 vs 50,265). During pretraining, RoBERTa adopts bidirectional objectives and novel strategies like larger training batches. It can be concluded that, as opposed to model size, \textbf{training objectives and strategies reflect the way 
that PLMs memorize information, making significant impacts on PLMs' memory ability}. Besides, we can clearly observe that all PLMs achieve their best results in T-REx (created from Wikipedia triples) among LAMA, and perform relatively well on Wikipedia. This implies that PLMs indeed remember a large proportion of knowledge and language patterns from pretraining corpora.

To test the memory efficiency, we fine-tune five models, BERT, ALBERT, GPT-2, BART, and XLNet, for several epochs. As shown in Figure~\ref{fig:memory-efficiency}, to achieve a reference performance, the bidirectional training objective like BERT needs fewer epochs than other kinds of objectives. This further implies that the bidirectional training objective is also helpful to facilitate the memory efficiency since bidirectional language modeling can make PLMs more quickly capture the language patterns.

Based on the memory test results, we further analyze how to effectively elicit the information from PLMs' memory. LAMA hand-crafts templates to test PLMs by filling the \texttt{[MASK]} token. Therefore, we conduct a pilot study on designing different templates for two relations in Google-RE. Table~\ref{tab:memory-prompt} shows that different templates can result in substantial differences in eliciting PLMs' memory. The bidirectional LMs, \eg BERT, show relatively adaptability to varying templates, further verifying their strength in memory ability. Therefore, with large-scale knowledge stored in PLMs, how to derive an effective and appropriate method to provoke them is a key challenge.

\begin{table*}[t]
	\small
	\centering
	\begin{tabular}{lcccccccccc}
		\toprule
		\multirow{2}{*}{\textbf{Models}}& \textbf{WNLI} & \textbf{CoLA} & \textbf{MNLI}      & \textbf{RTE}  & \textbf{QNLI} & \textbf{SST-2} & \textbf{QQP}       & \textbf{STS-B}     & \textbf{MRPC}      & \textbf{Avg.} \\ 
		& Acc. & Matt. & M./MM. & Acc. & Acc. & Acc. & F1/Acc. & P/S Corr. & F1/Acc. \\
		
		\midrule
		BERT\textsubscript{\textsc{base}}        & 65.1 & 52.1 & 84.6/83.4 & 66.4 & 90.5 & 93.5  & 69.9/88.2 & 77.4/73.7 & 79.0/85.1 & 76.5     \\
		BERT\textsubscript{\textsc{large}}       & 65.1 & 60.5 & 86.7/85.9 & 70.1 & 92.7 & 94.9  & 72.1/89.3 & 87.6/86.5 & 85.4/89.3 & 80.5     \\
		RoBERTa\textsubscript{\textsc{base}}     & 65.1 & 61.4 & 87.4/87.2 & 75.1 & 92.9 & 95.7  & 72.5/89.4 & 89.2/88.5 & 87.5/90.7 & 81.8     \\
		RoBERTa\textsubscript{\textsc{large}}    & \underline{89.0} & \underline{67.8} & \underline{90.8}/\underline{90.2} & \underline{88.2} & \underline{98.9} & \underline{96.7}  & \underline{74.3}/\underline{90.2} & \underline{92.2}/\underline{91.9} & \underline{89.9}/\underline{92.4} & \underline{88.5}     \\
		ALBERT\textsubscript{\textsc{xlarge}}    & 65.8 & 58.2 & 35.6/36.5 & 62.5 & 94.2 & 95.1  & 71.7/88.9 & 87.6/86.6 & 69.8/80.3 & 72.7     \\
		ALBERT\textsubscript{\textsc{xxlarge}}   & 64.4 & 64.7 & 89.7/89.6 & 70.4 & 95.3 & 96.0  & 70.7/88.4 & 91.3/90.6 & 68.1/80.4 & 80.6     \\
		GPT-2\textsubscript{\textsc{small}}       & 54.8 & 33.8 & 81.1/81.4 & 62.1 & 86.7 & 91.2  & 69.8/87.9 & 79.0/76.5 & 76.9/83.6 & 71.9     \\
		GPT-2\textsubscript{\textsc{medium}}      & 54.1 & 50.5 & 84.8/84.5 & 63.6 & 91.2 & 92.1  & 71.4/88.6 & 84.3/82.7 & 80.0/85.5 & 75.8     \\
		XLNet\textsubscript{\textsc{base}}       & 58.9 & 26.2 & 86.1/85.3 & 59.9 & 91.3 & 94.0  & 71.5/88.9 & 83.9/82.9 & 84.3/88.3 & 74.0     \\
		XLNet\textsubscript{\textsc{large}}      & \textbf{92.5} & \textbf{70.2} & \textbf{90.9}/\textbf{90.9} & \textbf{88.5} & \textbf{99.0} & \textbf{97.1}  & \textbf{74.7}/\textbf{90.4} & \textbf{93.0}/\textbf{92.6} & \textbf{90.5}/\textbf{92.9} & \textbf{89.5}     \\
		UniLM\textsubscript{\textsc{base}}       & 65.1 & 49.0 & 83.0/82.2 & 60.3 & 88.7 & 92.3  & 70.7/88.4 & 82.3/81.4 & 84.3/88.7 & 76.2     \\
		UniLM\textsubscript{\textsc{large}}      & 65.1 & 61.1 & 87.0/85.9 & 70.9 & 92.7 & 94.5  & 71.5/89.2 & 86.6/85.3 & 85.2/89.1 & 80.5     \\
		ERNIE\textsubscript{\textsc{base}}       & 65.1  & 52.3 & 84.0/83.2 & 68.8 & 91.3 & 93.5  & 70.5/88.4  & 85.1/83.8   & 80.3/85.9  & 70.7 \\
		T5\textsubscript{\textsc{base}}          & 78.8 & 51.1 & 87.1/86.2 & 80.1 & 93.7 & 95.2  & 72.6/89.4 & 89.4/88.6 & 87.5/90.7 & 82.7     \\
		T5\textsubscript{\textsc{large}}         & 85.6 & 61.2 & 89.9/89.6 & 87.2 & 94.8 & 96.3  & 73.9/89.9 & 89.9/89.2 & 89.8/92.4 & 86.4     \\
		BART\textsubscript{\textsc{base}}        & 65.1 & 52.8 & 85.1/84.3 & 69.5 & 92.6 & 94.4  & 72.5/89.7 & 87.6/86.6 & 86.1/89.5 & 79.5     \\
		BART\textsubscript{\textsc{large}}       & 58.9 & 62.4 & 90.2/89.3 & 83.5 & 94.8 & 96.3  & 73.6/90.1 & 91.1/90.4 & 87.8/91.1 & 83.1     \\
		ProphetNet\textsubscript{\textsc{large}} & 52.1 & 24.2 & 81.3/80.8 & 51.3 & 93.2 & 93.6  & 70.6/88.1 & 73.5/72.3 & 69.7/80.8 & 69.2     \\
		\bottomrule
	\end{tabular}
	\caption{Comprehension tests results on GLUE (test set). All results are scored by the GLUE evaluation server\protect\footnotemark. }
	\label{tab:comprehension-main}
\end{table*}
\footnotetext{\url{https://gluebenchmark.com/}}

\begin{figure}[t]
	\centering
	\subfigure[CoLA]{
		\centering
		\includegraphics[width=0.22\textwidth]{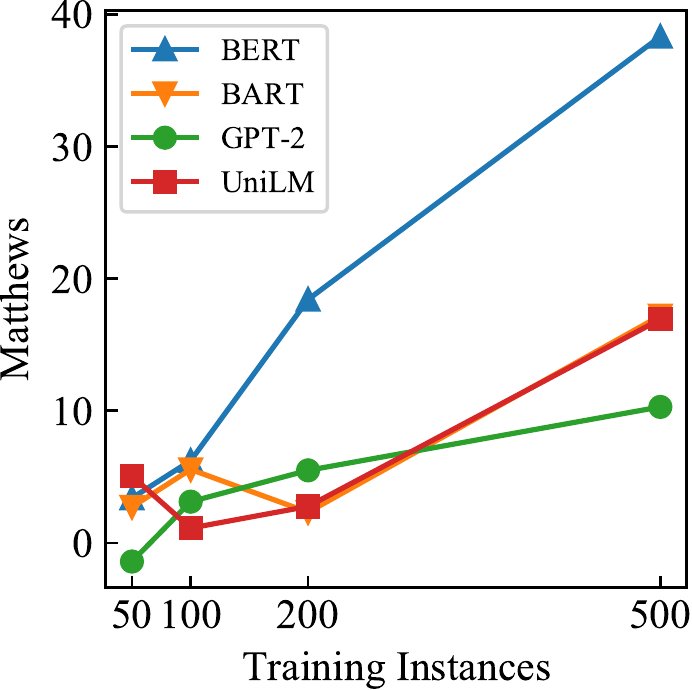}
	}
	\subfigure[QNLI]{
		\centering
		\includegraphics[width=0.22\textwidth]{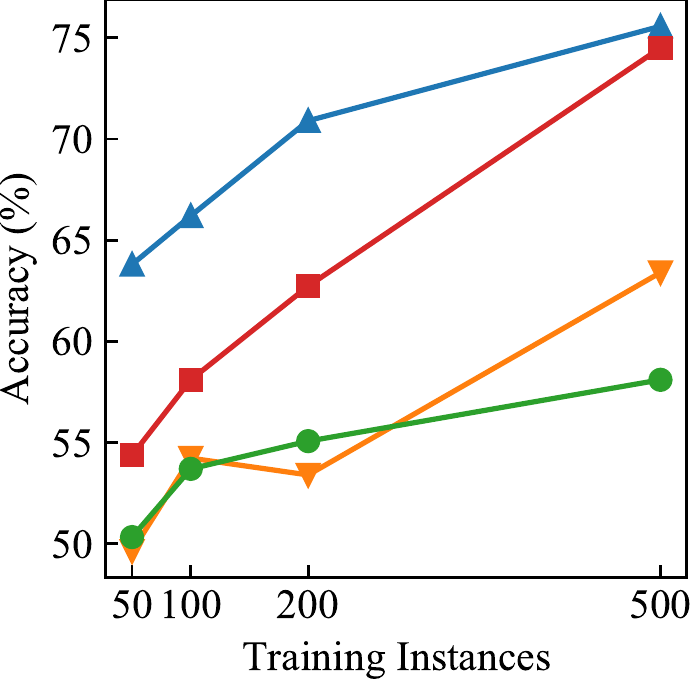}
	}
	\centering
	\caption{Few-shot results of four PLMs on CoLA and QNLI tasks.}
	\label{fig:comprehension-few-shot}
\end{figure}



\subsection{Comprehension Tests}
\label{sec-comprehension}

\paratitle{Datasets and Metrics.} In comprehension tests, we take into account three aspects of comprehension ability, including vocabulary, background knowledge, and linguistic structures. Therefore, we employ five datasets for comprehension tests, \ie GLUE~\cite{WangSMHLB19}, SuperGLUE~\cite{WangPNSMHLB19}, SQuAD v1.1~\cite{RajpurkarZLL16}, SQuAD v2.0~\cite{RajpurkarJL18}, and RACE~\cite{LaiXLYH17}. Among these datasets, GLUE and SuperGLUE are two widely-used comprehension benchmarks. Several tasks, like word sense disambiguation and coreference resolution, can assess PLMs' understanding of vocabulary meaning and grammatical structure of a text. By contrast, SQuAD v1.1\&v2.0, and RACE are three popular question answering datasets. To answer the natural language questions, PLMs should be aware of the background knowledge about some particular topic. For example, to answer the question ``\emph{what can be used as rewards for dog training?}'', the background knowledge ``\emph{dogs like bones}'' will be helpful for PLMs to answer ``\emph{bones}''. For evaluation, we report the corresponding metrics results for each task, such as the \textit{Matthews corr.} metric for CoLA.

\begin{figure}[t]
	\centering
	\includegraphics[width=0.45\textwidth]{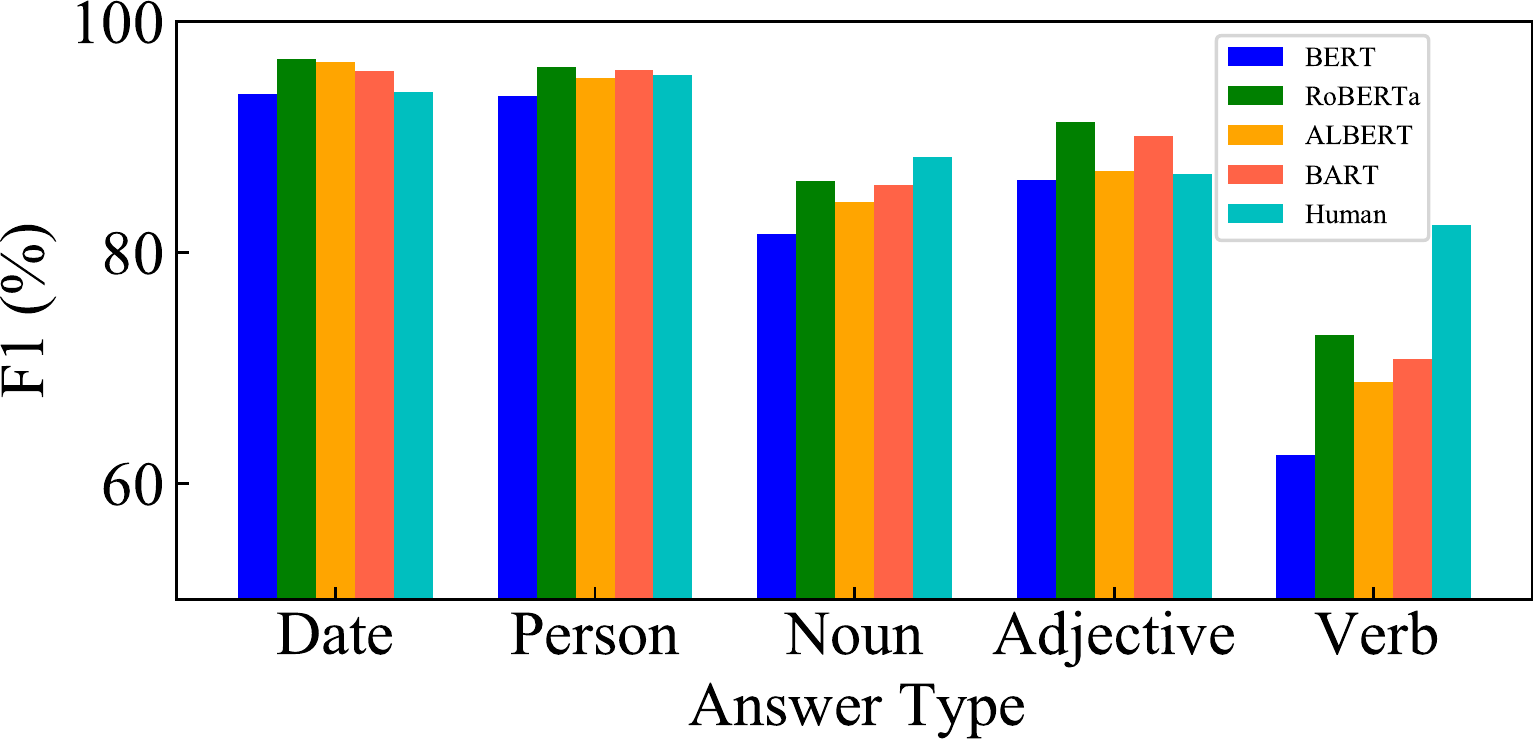}
	\caption{PLMs Performance on SQuAD v1.1\&v2.0 stratified by five types of answer.}
	\label{fig:comprehension-answer-type}
\end{figure}

\paratitle{Results and Analysis}. Table~\ref{tab:comprehension-main} presents the  results of comprehension test in GLUE dataset (results in other four datasets can be found in Appendix~\ref{comprehension}). The last column in this table indicates the average overall performance across all tasks. Interestingly, the models behaving well in memory tests (\eg RoBERTa and XLNet) also present good results in many comprehension tasks. The results indicate that \textbf{the improvement on memory ability is beneficial for the performance of comprehension ability}, which is in line with our intuition. Compared with bidirectional language modeling in BERT, permutation language modeling (relying on all permutations of the factorization order) used in XLNet enables PLMs to learn more context for enhancing PLMs' understanding of text, which seems to be effective for good comprehension ability.  

Among these tasks, we observe a significant performance drop in the linguistic acceptability task (CoLA) since it has different data distribution from the pretraining corpora~\cite{wang2021entailment}. This kind of sensitiveness to unfamiliar tasks is also reflected in Figure~\ref{fig:comprehension-few-shot}, where the model performance on CoLA shows a more volatile fluctuation (ranging from 10 to 35) than QNLI (ranging from 15 to 20). It indicates that \textbf{the performance of PLMs is closely related to the similarity of data distributions in pretraining and fine-tuning}. To solve this challenge, it will be better to adopt intermediate fine-tuning, which involves first fine-tuning PLMs on an intermediate dataset similar to the final target dataset and then transferring tuned PLMs to the final dataset.

\begin{table*}[t]
 \small
 \centering
 \begin{tabular}{lcccccccccc}
  \toprule
  		\multirow{2.5}*{\textbf{Datasets}}& \multicolumn{3}{c}{\textbf{Bidirectional}} & \multicolumn{1}{c}{\textbf{Uni.}}& \multicolumn{2}{c}{\textbf{Hybrid}} & \multicolumn{1}{c}{\textbf{KE}}  & \multicolumn{3}{c}{\textbf{Text-to-Text}}  \\
\cmidrule(lr){2-4}\cmidrule(lr){5-5}\cmidrule(lr){6-7}\cmidrule(lr){8-8}\cmidrule(lr){9-11}
& BERT          & RoBERTa       & ALBERT     & GPT-2 & XLNet  & UniLM    & ERNIE &   T5       & BART  & ProphetNet \\ 
\midrule
	\textbf{CQA}        & 55.9 & 72.2                & \textbf{80.0} & 60.8 & 62.9       & 62.3 & 54.1 & 69.8       & {\ul 75.8}    & 21.3 \\
	\textbf{ROCStories} & 90.2 & \textbf{97.4}       & {\ul 97.1}    & 59.9 & 93.8       & 86.9 & 84.7 & 91.4       & 91.7          & 82.2 \\
	\textbf{SWAG}       & 86.3 & {\ul 89.9} & \textbf{90.7} & 79.7 & 86.8       & 83.1 &   80.2   & 73.7       & 87.9          & 70.1 \\
	\textbf{HellaSwag}  & 47.3 & {\ul 85.2}          & \textbf{90.1} & 60.4 & 79.7       & 46.7 &   44.5   & 79.1       & 76.6          & 26.4 \\
	\textbf{SM-A}       & 89.4 & \textbf{93.0}       & 92.5          & 88.7 & 83.7       & 89.3 & 88.7 & {\ul 92.7} & 82.9          & 85.5 \\
	\textbf{SM-B}       & 85.8 & \textbf{92.3}       & \textbf{92.3} & 73.4 & {\ul 88.7}       & 86.4 &   87.7   & 88.2       & 67.9          & 78.0 \\
	\textbf{ARCT}       & 71.2 & 57.9                & 79.5          & 66.7 & {\ul 83.1} & 72.3 & 73.7 & 69.4       & \textbf{84.2} & 65.5 \\
  \bottomrule
 \end{tabular}
 \caption{Reasoning tests results on seven datasets (test set). CQA is short for CommonsenseQA. SM-A and SM-B denote the Task A and Task B of Sense Making, respectively. We report the results of \textsc{Large} version for each model in this table and more results can be found in the Appendix~\ref{reasoning}.}
 \label{tab:reasoning-main}
\end{table*}


To gain more insights into PLMs' comprehension ability, we choose four representative PLMs (\ie BERT, RoBERTa, ALBERT, and BART) and humans to analyze their performance across the answer types of SQuAD v1.1\&v2.0. The results in Figure~\ref{fig:comprehension-answer-type} show that PLMs perform well on simple answers such as dates and persons. For these categories of answers, there are usually only a few plausible candidates and most answers are single tokens. The models are more challenged on other intricate answer types (\eg noun and verb phrases) because there are many more plausible candidates and multiple tokens. Thus, improving PLMs' understanding of intricate named entities during the pretraining phase will possibly benefit PLMs' comprehension ability later.

\begin{figure}[t]
	\centering
	\subfigure[BERT\textsubscript{\textsc{large}}]{
		\centering
		\includegraphics[width=0.22\textwidth]{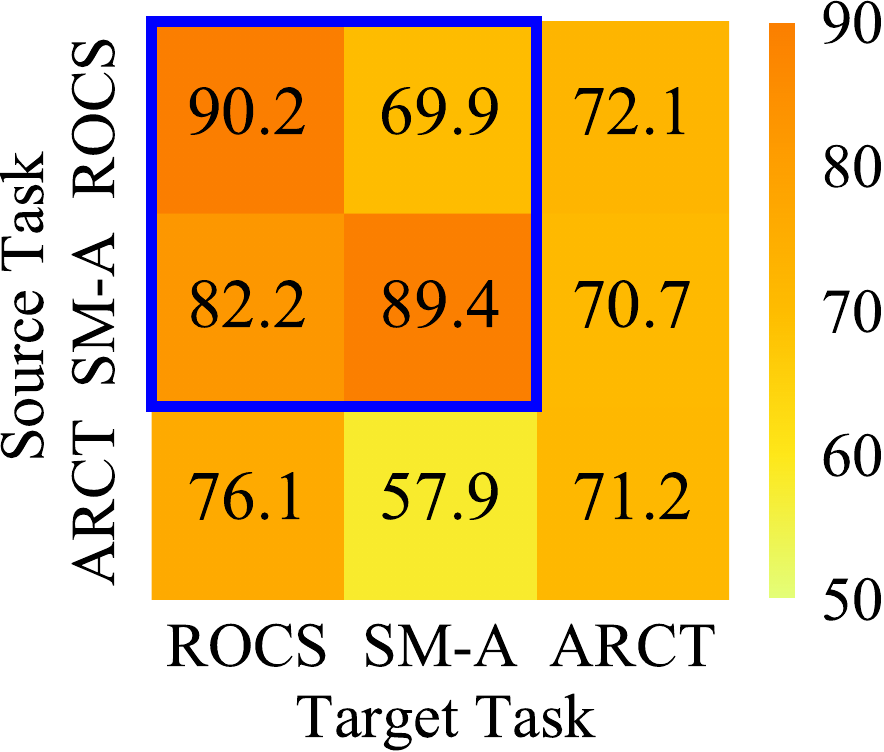}
	}
	\subfigure[T5\textsubscript{\textsc{large}}]{
		\centering
		\includegraphics[width=0.22\textwidth]{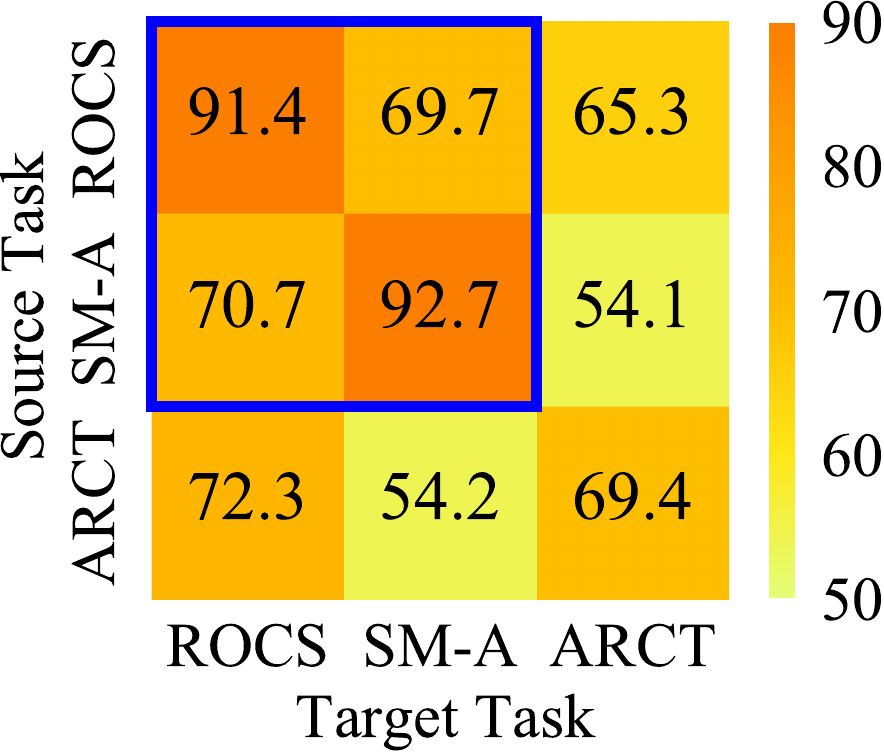}
	}
	\centering
	\caption{Heatmaps of two-stage transfer learning.}
	\label{fig:reasoning-intermediate}
	\vspace{-0.2cm}
\end{figure}

\subsection{Reasoning Tests}
\label{sec-reasoning}

\paratitle{Datasets and Metrics.} In reasoning tests, we mainly consider three forms of reasoning, \ie commonsense reasoning, deductive reasoning, and abductive reasoning, focusing on commonsense utilization, conclusion induction, and reason derivation, respectively. For evaluation, we choose six reasoning datasets, namely CommonsenseQA~\cite{TalmorHLB19}, ROCStories~\cite{mostafazadeh2016corpus}, SWAG~\cite{ZellersBSC18}, HellaSwag~\cite{ZellersHBFC19}, Sense Making~\cite{WangLZLG19}, and ARCT~\cite{HabernalWGS18a}. Specifically, CommonsenseQA requires PLMs to reason about commonsense knowledge in human experience of everyday life~\cite{liu2004conceptnet}. ROCStories, SWAG, HellaSwag, and Sense Making Task A are concerned with deriving the conclusions of stories and events, while Sense Making Task B and ARCT focus on identifying the reason behind a statement. For evaluation, we report the \textit{Accuracy} results for each dataset.

\paratitle{Results and Analysis}. Table~\ref{tab:reasoning-main} shows the model performances in reasoning ability. It can be clearly observed that performing well in comprehension tests, ALBERT and RoBERTa also achieve stronger performance in almost all reasoning tasks. In pretraining, ALBERT introduces an inter-sentence coherence objective to capture the relationship among sentences, which is helpful for the sentence-level reasoning ability of PLMs. It has been found that the next sentence prediction (NSP) loss in BERT might hurt the performance of PLMs in sentence-level tasks of downstream datasets~\cite{liu2019roberta}. Interestingly, despite being the best in comprehension tests, XLNet does not perform as well as we expected in reasoning tests. We speculate that the permutation operation in XLNet disturbs the semantic relationship between sentences, thus leading to poor reasoning ability. 
\textbf{To improve PLMs' reasoning ability, it would be useful to design sentence-level reasoning objectives like inter-sentence coherence loss in ALBERT}.
Moreover, despite incorporating knowledge, ERNIE still shows mediocre performance in knowledge-related datasets such as CQA. A possible reason might be that ERNIE only uses trained KB embeddings to enhance semantic representations but ignores the reasoning structure of KBs. This inspires us that designing appropriate and effective fusion methods to integrate knowledge is more important.

To further analyze the transferability of PLMs' reasoning ability, we conduct a two-stage study on three task datasets, \ie ROCStories, SM-A, and ARCT. We first train PLMs on source tasks with full data and then fine-tune PLMs on target tasks with ten instances. In Figure~\ref{fig:reasoning-intermediate}, it can be observed that \textbf{PLMs have better reasoning transferability between similar tasks} such as deductive reasoning tasks (ROCStories and SM-A). This shows that model performance on data-scarce reasoning tasks can be improved by incorporating additional training on data-rich similar tasks~\cite{wang2021entailment}.


\begin{table*}[t]
	\small
	\centering
	\begin{tabular}{lrrrrrrrrrrrr}
		\toprule
		\multirow{2.5}*{\textbf{Models}} & \multicolumn{3}{c}{\textbf{CNN/DailyMail}} & \multicolumn{3}{c}{\textbf{GigaWord}} & \multicolumn{3}{c}{\textbf{SQuAD}} & \multicolumn{3}{c}{\textbf{WritingPrompts}} \\
		\cmidrule(r){2-4}\cmidrule(r){5-7}\cmidrule(r){8-10}\cmidrule(r){11-13}
		 & \makecell[c]{R-1} & \makecell[c]{R-2} & \makecell[c]{R-L} & \makecell[c]{R-1} & \makecell[c]{R-2} & \makecell[c]{R-L} & \makecell[c]{B-4} & \makecell[c]{R-L} & \makecell[c]{ME} & \makecell[c]{B-4} & \makecell[c]{R-L} & \makecell[c]{ME} \\
		\midrule
		GPT-2      &      27.00     &     8.00    &    23.08    & 23.72          & 8.12           & 21.56          & 8.48           & 18.82          & 26.77          & 14.47          & {\ul 3.23}    & {\ul 7.29}    \\
		UniLm    & 43.44          & 20.21          & 40.51          & 38.45          & 19.45          & 35.75          & 4.42           & 17.43          & 20.13          & \textbf{26.88} & 1.84          & 5.01          \\
		T5        & 42.50          & 20.68          & 39.75          & 34.75          & 16.26          & 31.49          & 11.19          & 22.35          & 30.53          & 8.61           & \textbf{4.19} & \textbf{9.51} \\
		BART       & {\ul 44.16}    & \textbf{21.28} & {\ul 40.90}    & {\ul 39.41}    & {\ul 20.21}    & {\ul 36.42}    & \textbf{15.87} & \textbf{25.47} & \textbf{38.42} & 14.72          & 3.14          & 7.08          \\
		ProphetNet & \textbf{44.20} & {\ul 21.17}    & \textbf{41.30} & \textbf{39.51} & \textbf{20.42} & \textbf{36.69} & {\ul 14.20}    & {\ul 23.97}    & {\ul 35.99}    & {\ul 19.31}    & 2.59          & 7.19         \\
		\bottomrule
	\end{tabular}
	\caption{Composition tests results on four datasets. R-1, R-2, R-L are short for ROUGE-1, ROUGE-2, ROUGE-L respectively. B-4 and MT denote BLEU-4 and METEOR, respectively. We report the result of \textsc{Large} version for each model in this table and more results can be found in the Appendix~\ref{composition}.}
	\label{tab:composition-main}
\end{table*}

\begin{table}[t]
	\small
	\centering
	\begin{tabular}{lr|rrrr}
		\toprule
		 \multirow{2.5}*{\textbf{Models}} & \multicolumn{5}{c}{\textbf{GigaWord}} \\
		 \cmidrule{2-6}
		 & \textbf{TT (\%)} & \textbf{Flu.} & \textbf{Info.} & \textbf{Acc.} & \textbf{Overall} \\
		\midrule
		GPT-2       & 26.09          & 3.11          & 2.79          & 2.64          & 4.87          \\
		UniLM      & 50.34          & \textbf{4.02} & {\ul 3.49}    & 3.45          & {\ul 6.73}    \\
		T5         & \textbf{53.67} & 3.95          & 3.45          & {\ul 3.46}    & 6.68          \\
		BART       & 51.10          & 4.01          & 3.46          & \textbf{3.49} & {\ul 6.73}    \\
		ProphetNet & {\ul 53.02}    & {\ul 3.99}    & \textbf{3.52} & 3.45          & \textbf{6.74} \\
		\cmidrule{1-6}
		Gold       & 40.77          & 3.61          & 3.29          & 3.15          & 6.05         \\
		\midrule
		\multirow{2.5}*{\textbf{Models}} & \multicolumn{5}{c}{\textbf{WritingPrompts}} \\
		\cmidrule{2-6}
		& \textbf{TT (\%)} & \textbf{Flu.} & \textbf{Info.} & \textbf{Rel.} & \textbf{Overall} \\
		\midrule
		GPT-2 & \textbf{45.70} & \textbf{3.42} & \textbf{3.17} & {\ul 3.20}    & {\ul 5.87}    \\
  		UniLM & 1.20           & 1.32          & 1.88          & 2.03          & 2.74          \\
  		T5 & 34.40          & 3.01          & 2.80          & 3.09          & 5.18          \\
  		BART & {\ul 45.20}    & 3.37          & {\ul 3.16}    & \textbf{3.39} & \textbf{5.96} \\
  		ProphetNet & 29.60          & 2.95          & 2.91          & 3.10          & 5.18          \\
  		\cmidrule{1-6}
  		Gold & 71.30          & 3.79          & 4.07          & 3.87          & 7.37           \\
		\bottomrule
	\end{tabular}
	\caption{Turing test (TT) and human scores on the test set of GigaWord and WritingPrompts. Flu., Info., Acc. and Rel. denote fluency, informativeness, accuracy and relevance respectively. We report the result of \textsc{Large} version for each model in this table and more results can be found in the Appendix~\ref{composition}.}
	\label{tab:turing-test}
\end{table}

\begin{figure}[t]
	\centering
	\subfigure[GigaWord]{
		\centering
		\includegraphics[width=0.22\textwidth]{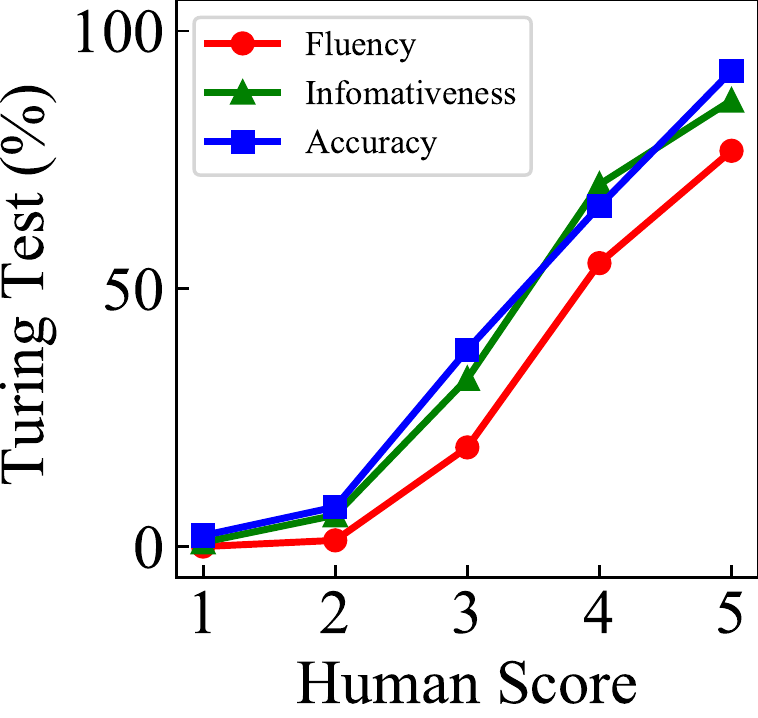}
	}
	\subfigure[WritingPrompts]{
		\centering
		\includegraphics[width=0.22\textwidth]{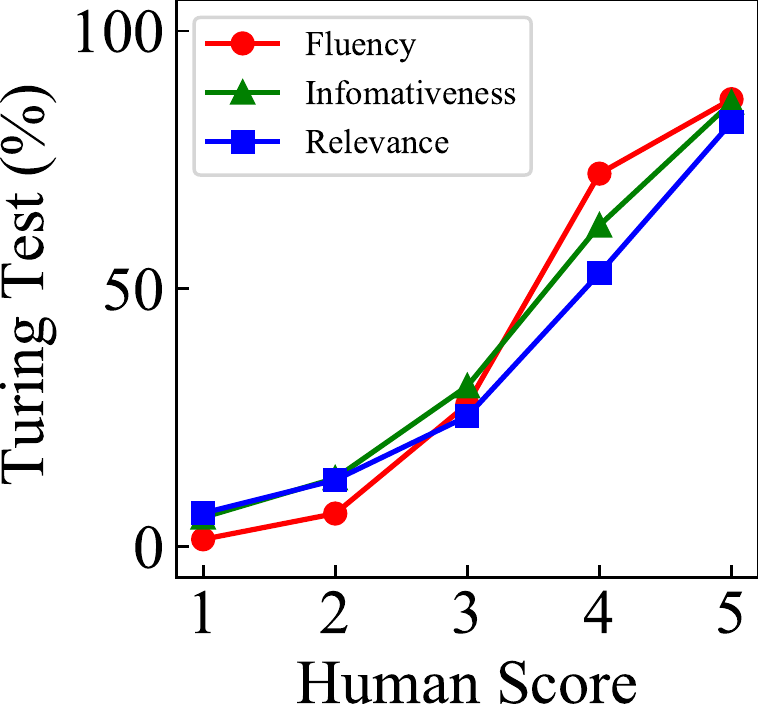}
	}
	\centering
	\caption{Impact factors of Turing Test.}
	\label{fig:composition-analysis}
	\vspace{-0.3cm}
\end{figure} 

\subsection{Composition Tests}

\paratitle{Datasets and Metrics.} Composition is similar to the text generation task, aiming at generating new content from scratch. Therefore, we use four text generation benchmarks for composition tests, \ie WritingPrompts~\cite{LewisDF18} on story generation, CNN/Daily Mail~\cite{HermannKGEKSB15} and GigaWord~\cite{RushCW15} on text summarization, and SQuAD v1.1~\cite{RajpurkarZLL16} on question generation. According to the length of the target text, text summarization and question generation is short text generation, while story generation is long text generation. For evaluation, we adopt three automatic metrics, \ie BLEU~\cite{papineni2002bleu}, ROUGE~\cite{lin2004rouge}, and METEOR~\cite{BanerjeeL05}. 
Besides, following~\cite{zou2021controllable}, we conduct human test from five aspects, \ie \textit{Fluency}, \textit{Informativeness}, \textit{Accuracy}, \textit{Relevance} and \textit{Overall}. 
The overall score is rated from 1 to 10, while the others are rated from 1 to 5. Inspired by~\citet{turing2009computing}, we further design a Turing test to assess the generation ability of PLMs, where a human interrogator is requested to distinguish whether the given text is generated by a human. From the generated texts of each model and gold texts, we randomly select 500 texts scored by judges.  More details of human test and Turing test are shown in Appendix~\ref{composition}.

%
%
%
%

\paratitle{Results and Analysis}. Table~\ref{tab:composition-main} and Table~\ref{tab:turing-test} present the automatic evaluation and human evaluation results on composition ability, respectively. We can observe that ProphetNet and BART achieve great performance on short text generation, while GPT-2 and T5 show better results on long text generation. Specifically, BART  employs denoising objectives for reconstructing the corrupted original text, and ProphetNet adopts future $n$-gram prediction, which is flexible for modeling the semantic relations between tokens and phrases in short texts. However, in long texts, a small ratio of masked tokens (\ie 15\%) might be not effective in capturing the complex long-range dependency. By comparison, the left-to-right prediction objective in GPT-2 can be more suitable to model the long-range semantic continuity in long texts, and T5 has the largest model size to achieve a strong composition ability. For composition ability, we conclude that \textbf{the denoising objective is helpful for short text composition, while the left-to-right objective is more powerful for long text composition}. Besides, the model size is also an important factor in improving PLMs' composition ability.

To further investigate what factors affect the pass rate of the Turing test, we deeply analyze the intermediate scoring results in the human test and Turing test. As shown in Figure~\ref{fig:composition-analysis}, we calculate the pass rate of the Turing test for each human test metric across 1 to 5 scale. Moreover, we compute the Pearson correlation coefficient between the pass rate and each metric. In story generation (WritingPrompts), the coefficients for \textit{Fluency}, \textit{Informativeness}, and \textit{Relevance} are 96.63, 97.93, 96.44, respectively. While, in text summarization (GigaWord), the coefficients for \textit{Fluency}, \textit{Informativeness}, and \textit{Accuracy} are 96.08, 97.67, 98.38, respectively. From these analysis results, we can conclude that \textit{Informativeness} is more important for story generation, while \textit{Accuracy} is more influential in text summarization. Besides, we compute the text similarity between the generated texts from different PLMs, which is shown in Appendix~\ref{composition}.



\section{Discussion}
\label{sec-guide}

Based on the above four ability tests, we intend to provide a guideline for helping researchers choose, apply, interpret and design PLMs for NLP tasks.

In section~\ref{sec-comprehension}, we observe that the improvement in memory ability is likely to be helpful for the performance of comprehension ability. Hence, designing PLMs with special objectives like bidirectional language modeling in BERT and strategies like larger training batches in RoBERTa for larger memory capacity will further benefit PLMs in the downstream comprehension tasks. Besides, when applying PLMs to downstream tasks, the similarity of data distribution between pretraining and fine-tuning has a great impact on PLMs performance. Possible solutions such as introducing intermediate tasks or datasets can alleviate such a discrepancy. Moreover, we further find some limitations in PLMs' comprehension ability, where PLMs are good at simple single-token answer types in QA such as dates but perform worse in complex phrase answers.

Compared to comprehension, reasoning in section~\ref{sec-reasoning} is much more intricate and usually involves inferring the semantic relationships among multiple sentences. Therefore, PLMs such as ALBERT trained with sentence-level objectives can be more suitable for conducting reasoning tasks. Intuitively, incorporating sentence-level objectives during pretraining will help PLMs learn the correlation among different sentences. Note that PLMs have better reasoning transferability between similar tasks, thus data-scarce reasoning tasks can be improved by first training on data-rich tasks.

For composition ability, PLMs with denoising training objectives perform much better on short text composition, while PLMs with left-to-right objectives or larger model size are more suitable for long text composition. This might be because PLMs with different training objectives can finally capture different ranges of semantic dependency between tokens and phrases. Moreover, to obtain a higher pass rate of Turing test, different text generation tasks will be concerned with varying factors, such as informativeness is much more critical for story generation.

\section{Related Work}

\paratitle{Pretrained Language Models}. Owing to the great achievements Transformer~\cite{VaswaniSPUJGKP17} has made, the paradigm of pretrained language models (PLMs) is thriving~\cite{radford2019language,DevlinCLT19,liu2019roberta,LewisLGGMLSZ20,RaffelSRLNMZLL20}. It is widely recognized that PLMs can learn massive knowledge from corpora~\cite{LiTZW21}, leading to significant progress in various language tasks~\cite{textbox,LiTZWYW21}. With such encouraging results in extensive NLP tasks, it is a non-trivial topic to systematically evaluate the abilities of PLMs, which can further deepen our understanding of PLMs and facilitate their application to more fields. 

\paratitle{Language Model Evaluation}. Many efforts have studied the evaluation of language model performance. \citet{Liu0BPS19} evaluate BERT~\cite{DevlinCLT19}, GPT~\cite{radford2018improving}, and ELMo~\cite{PetersNIGCLZ18} on a variety of linguistics tasks. Their findings indicate that the features generated by PLMs are sufficient for good performance on a board set of tasks but fall short on tasks requiring fine-grained linguistic knowledge. \citet{tenney2019you} evaluate similar models on a range of sub-sentence linguistic analysis tasks, showing that PLMs encode both syntax and semantics into parameters. \citet{ZhouZCH20} also report that PLMs can learn rich knowledge but focus on evaluating the commonsense. However, these studies only look at one dimension of PLMs ability evaluation. Other work such GLUE~\cite{WangSMHLB19} and CLUE~\cite{CLUEbenchmark} just consider a simple mixture of multiple tasks lacking comprehensive evaluation. To the best of our knowledge, this is the first work to systematically evaluate PLMs by defining various kinds of language abilities and performing extensive comparison.
\section{Conclusion}

This paper investigates the general language ability evaluation of pretrained language models. We design four kinds of language abilities of PLMs, including memory, comprehension, reasoning, and composition, and measure ten widely-used PLMs within five categories. For each language ability, we select multiple representative tasks to quantitatively evaluate the performance of PLMs. Our experimental results demonstrate that PLMs with varying objectives and strategies are good at different ability tests. Note that our final predicted outputs of PLMs can also be reused as an open resource for more depth and granularity in analyzing PLMs' language abilities. As a result, it is believed that this study will benefit future work about choosing or designing suitable PLMs for the target NLP tasks based on their properties.

\section*{Acknowledgement}

This work was partially supported by Beijing Natural Science Foundation under Grant No. 4222027,  National Natural Science Foundation of China under Grant No. 61872369 and 82161148011, Beijing Outstanding Young Scientist Program under Grant No. BJJWZYJH012019100020098, and the Outstanding Innovative Talents Cultivation Funded Programs 2021 of Renmin University of China.
We are grateful to Amazon Web Services for providing efficient GPU computing resource support and technical support for this NLP research project. Xin Zhao is the corresponding author.

\bibliography{anthology}

\begin{thebibliography}{56}
\expandafter\ifx\csname natexlab\endcsname\relax\def\natexlab#1{#1}\fi

\bibitem[{Banerjee and Lavie(2005)}]{BanerjeeL05}
Satanjeev Banerjee and Alon Lavie. 2005.
\newblock {METEOR:} an automatic metric for {MT} evaluation with improved
  correlation with human judgments.
\newblock In \emph{Proceedings of the Workshop on Intrinsic and Extrinsic
  Evaluation Measures for Machine Translation and/or Summarization@ACL 2005,
  Ann Arbor, Michigan, USA, June 29, 2005}, pages 65--72. Association for
  Computational Linguistics.

\bibitem[{Berninger(1999)}]{berninger1999coordinating}
Virginia~W Berninger. 1999.
\newblock Coordinating transcription and text generation in working memory
  during composing: Automatic and constructive processes.
\newblock \emph{Learning Disability Quarterly}, 22(2):99--112.

\bibitem[{Brown et~al.(2020)Brown, Mann, Ryder, Subbiah, Kaplan, Dhariwal,
  Neelakantan, Shyam, Sastry, Askell et~al.}]{brown2020language}
Tom~B Brown, Benjamin Mann, Nick Ryder, Melanie Subbiah, Jared Kaplan, Prafulla
  Dhariwal, Arvind Neelakantan, Pranav Shyam, Girish Sastry, Amanda Askell,
  et~al. 2020.
\newblock Language models are few-shot learners.
\newblock \emph{arXiv preprint arXiv:2005.14165}.

\bibitem[{Cain and Oakhill(2008)}]{cain2008children}
Kate Cain and Jane Oakhill. 2008.
\newblock \emph{Children's comprehension problems in oral and written language:
  A cognitive perspective}.
\newblock Guilford Press.

\bibitem[{Connors(1997)}]{connors1997composition}
Robert Connors. 1997.
\newblock \emph{Composition-rhetoric: Backgrounds, theory, and pedagogy}.
\newblock University of Pittsburgh Pre.

\bibitem[{Devlin et~al.(2019)Devlin, Chang, Lee, and Toutanova}]{DevlinCLT19}
Jacob Devlin, Ming{-}Wei Chang, Kenton Lee, and Kristina Toutanova. 2019.
\newblock \href {https://doi.org/10.18653/v1/n19-1423} {{BERT:} pre-training of
  deep bidirectional transformers for language understanding}.
\newblock In \emph{Proceedings of the 2019 Conference of the North American
  Chapter of the Association for Computational Linguistics: Human Language
  Technologies, {NAACL-HLT} 2019, Minneapolis, MN, USA, June 2-7, 2019, Volume
  1 (Long and Short Papers)}, pages 4171--4186. Association for Computational
  Linguistics.

\bibitem[{Dong et~al.(2019)Dong, Yang, Wang, Wei, Liu, Wang, Gao, Zhou, and
  Hon}]{00040WWLWGZH19}
Li~Dong, Nan Yang, Wenhui Wang, Furu Wei, Xiaodong Liu, Yu~Wang, Jianfeng Gao,
  Ming Zhou, and Hsiao{-}Wuen Hon. 2019.
\newblock \href
  {https://proceedings.neurips.cc/paper/2019/hash/c20bb2d9a50d5ac1f713f8b34d9aac5a-Abstract.html}
  {Unified language model pre-training for natural language understanding and
  generation}.
\newblock In \emph{Advances in Neural Information Processing Systems 32: Annual
  Conference on Neural Information Processing Systems 2019, NeurIPS 2019,
  December 8-14, 2019, Vancouver, BC, Canada}, pages 13042--13054.

\bibitem[{F.~Petroni and Riedel(2019)}]{petroni2019language}
A.~H. Miller P. Lewis A. Bakhtin Y.~Wu F.~Petroni, T.~Rockt{\"{a}}schel and
  S.~Riedel. 2019.
\newblock Language models as knowledge bases?
\newblock In \emph{In: Proceedings of the 2019 Conference on Empirical Methods
  in Natural Language Processing (EMNLP), 2019}.

\bibitem[{Fan et~al.(2018)Fan, Lewis, and Dauphin}]{LewisDF18}
Angela Fan, Mike Lewis, and Yann~N. Dauphin. 2018.
\newblock \href {https://www.aclweb.org/anthology/P18-1082/} {Hierarchical
  neural story generation}.
\newblock In \emph{Proceedings of the 56th Annual Meeting of the Association
  for Computational Linguistics, {ACL} 2018, Melbourne, Australia, July 15-20,
  2018, Volume 1: Long Papers}, pages 889--898. Association for Computational
  Linguistics.

\bibitem[{Habernal et~al.(2018)Habernal, Wachsmuth, Gurevych, and
  Stein}]{HabernalWGS18a}
Ivan Habernal, Henning Wachsmuth, Iryna Gurevych, and Benno Stein. 2018.
\newblock \href {https://doi.org/10.18653/v1/n18-1175} {The argument reasoning
  comprehension task: Identification and reconstruction of implicit warrants}.
\newblock In \emph{Proceedings of the 2018 Conference of the North American
  Chapter of the Association for Computational Linguistics: Human Language
  Technologies, {NAACL-HLT} 2018, New Orleans, Louisiana, USA, June 1-6, 2018,
  Volume 1 (Long Papers)}, pages 1930--1940. Association for Computational
  Linguistics.

\bibitem[{Hermann et~al.(2015)Hermann, Kocisk{\'{y}}, Grefenstette, Espeholt,
  Kay, Suleyman, and Blunsom}]{HermannKGEKSB15}
Karl~Moritz Hermann, Tom{\'{a}}s Kocisk{\'{y}}, Edward Grefenstette, Lasse
  Espeholt, Will Kay, Mustafa Suleyman, and Phil Blunsom. 2015.
\newblock \href
  {https://proceedings.neurips.cc/paper/2015/hash/afdec7005cc9f14302cd0474fd0f3c96-Abstract.html}
  {Teaching machines to read and comprehend}.
\newblock In \emph{Advances in Neural Information Processing Systems 28: Annual
  Conference on Neural Information Processing Systems 2015, December 7-12,
  2015, Montreal, Quebec, Canada}, pages 1693--1701.

\bibitem[{Johnson-Laird(1999)}]{johnson1999deductive}
Philip~N Johnson-Laird. 1999.
\newblock Deductive reasoning.
\newblock \emph{Annual review of psychology}, 50(1):109--135.

\bibitem[{Kaufman and Lichtenberger(2005)}]{kaufman2005assessing}
Alan~S Kaufman and Elizabeth~O Lichtenberger. 2005.
\newblock \emph{Assessing adolescent and adult intelligence}.
\newblock John Wiley \& Sons.

\bibitem[{Kyllonen and Christal(1990)}]{kyllonen1990reasoning}
Patrick~C Kyllonen and Raymond~E Christal. 1990.
\newblock Reasoning ability is (little more than) working-memory capacity?!
\newblock \emph{Intelligence}, 14(4):389--433.

\bibitem[{Lai et~al.(2017)Lai, Xie, Liu, Yang, and Hovy}]{LaiXLYH17}
Guokun Lai, Qizhe Xie, Hanxiao Liu, Yiming Yang, and Eduard~H. Hovy. 2017.
\newblock \href {https://doi.org/10.18653/v1/d17-1082} {{RACE:} large-scale
  reading comprehension dataset from examinations}.
\newblock In \emph{Proceedings of the 2017 Conference on Empirical Methods in
  Natural Language Processing, {EMNLP} 2017, Copenhagen, Denmark, September
  9-11, 2017}, pages 785--794. Association for Computational Linguistics.

\bibitem[{Lan et~al.(2020)Lan, Chen, Goodman, Gimpel, Sharma, and
  Soricut}]{LanCGGSS20}
Zhenzhong Lan, Mingda Chen, Sebastian Goodman, Kevin Gimpel, Piyush Sharma, and
  Radu Soricut. 2020.
\newblock \href {https://openreview.net/forum?id=H1eA7AEtvS} {{ALBERT:} {A}
  lite {BERT} for self-supervised learning of language representations}.
\newblock In \emph{8th International Conference on Learning Representations,
  {ICLR} 2020, Addis Ababa, Ethiopia, April 26-30, 2020}. OpenReview.net.

\bibitem[{Lewis et~al.(2020)Lewis, Liu, Goyal, Ghazvininejad, Mohamed, Levy,
  Stoyanov, and Zettlemoyer}]{LewisLGGMLSZ20}
Mike Lewis, Yinhan Liu, Naman Goyal, Marjan Ghazvininejad, Abdelrahman Mohamed,
  Omer Levy, Veselin Stoyanov, and Luke Zettlemoyer. 2020.
\newblock \href {https://doi.org/10.18653/v1/2020.acl-main.703} {{BART:}
  denoising sequence-to-sequence pre-training for natural language generation,
  translation, and comprehension}.
\newblock In \emph{Proceedings of the 58th Annual Meeting of the Association
  for Computational Linguistics, {ACL} 2020, Online, July 5-10, 2020}, pages
  7871--7880. Association for Computational Linguistics.

\bibitem[{Li et~al.(2021{\natexlab{a}})Li, Tang, He, Jiang, Hu, Xie, Chen, Yu,
  Zhao, and Wen}]{textbox}
Junyi Li, Tianyi Tang, Gaole He, Jinhao Jiang, Xiaoxuan Hu, Puzhao Xie, Zhipeng
  Chen, Zhuohao Yu, Wayne~Xin Zhao, and Ji-Rong Wen. 2021{\natexlab{a}}.
\newblock Textbox: A unified, modularized, and extensible framework for text
  generation.
\newblock In \emph{Proceedings of the 59th Annual Meeting of the Association
  for Computational Linguistics and the 11th International Joint Conference on
  Natural Language Processing: System Demonstrations}, pages 30--39.

\bibitem[{Li et~al.(2021{\natexlab{b}})Li, Tang, Zhao, Wei, Yuan, and
  Wen}]{LiTZWYW21}
Junyi Li, Tianyi Tang, Wayne~Xin Zhao, Zhicheng Wei, Nicholas~Jing Yuan, and
  Ji{-}Rong Wen. 2021{\natexlab{b}}.
\newblock Few-shot knowledge graph-to-text generation with pretrained language
  models.
\newblock In \emph{Findings of the Association for Computational Linguistics:
  {ACL/IJCNLP} 2021, Online Event, August 1-6, 2021}, volume {ACL/IJCNLP} 2021
  of \emph{Findings of {ACL}}, pages 1558--1568. Association for Computational
  Linguistics.

\bibitem[{Li et~al.(2021{\natexlab{c}})Li, Tang, Zhao, and Wen}]{LiTZW21}
Junyi Li, Tianyi Tang, Wayne~Xin Zhao, and Ji{-}Rong Wen. 2021{\natexlab{c}}.
\newblock Pretrained language model for text generation: {A} survey.
\newblock In \emph{Proceedings of the Thirtieth International Joint Conference
  on Artificial Intelligence, {IJCAI} 2021, Virtual Event / Montreal, Canada,
  19-27 August 2021}, pages 4492--4499. ijcai.org.

\bibitem[{Liang~Xu(2020)}]{CLUEbenchmark}
Lu~Li Hai Hu Chenjie Cao Weitang Liu Junyi Li Yudong Li Kai Sun Yechen Xu
  Yiming Cui Cong Yu Qianqian Dong Yin Tian Dian Yu Bo Shi Jun Zeng Rongzhao
  Wang Weijian Xie Yanting Li Yina Patterson Zuoyu Tian Yiwen Zhang He Zhou
  Shaoweihua Liu Qipeng Zhao Cong Yue Xinrui Zhang Zhengliang Yang
  Zhenzhong~Lan Liang~Xu, Xuanwei~Zhang. 2020.
\newblock Clue: A chinese language understanding evaluation benchmark.
\newblock \emph{arXiv preprint arXiv:2004.05986}.

\bibitem[{Lin(2004)}]{lin2004rouge}
Chin-Yew Lin. 2004.
\newblock Rouge: A package for automatic evaluation of summaries.
\newblock In \emph{Text summarization branches out}, pages 74--81.

\bibitem[{Liu and Singh(2004)}]{liu2004conceptnet}
Hugo Liu and Push Singh. 2004.
\newblock Conceptnet—a practical commonsense reasoning tool-kit.
\newblock \emph{BT technology journal}, 22(4):211--226.

\bibitem[{Liu et~al.(2019{\natexlab{a}})Liu, Gardner, Belinkov, Peters, and
  Smith}]{Liu0BPS19}
Nelson~F. Liu, Matt Gardner, Yonatan Belinkov, Matthew~E. Peters, and Noah~A.
  Smith. 2019{\natexlab{a}}.
\newblock \href {https://doi.org/10.18653/v1/n19-1112} {Linguistic knowledge
  and transferability of contextual representations}.
\newblock In \emph{Proceedings of the 2019 Conference of the North American
  Chapter of the Association for Computational Linguistics: Human Language
  Technologies, {NAACL-HLT} 2019, Minneapolis, MN, USA, June 2-7, 2019, Volume
  1 (Long and Short Papers)}, pages 1073--1094. Association for Computational
  Linguistics.

\bibitem[{Liu et~al.(2019{\natexlab{b}})Liu, Ott, Goyal, Du, Joshi, Chen, Levy,
  Lewis, Zettlemoyer, and Stoyanov}]{liu2019roberta}
Yinhan Liu, Myle Ott, Naman Goyal, Jingfei Du, Mandar Joshi, Danqi Chen, Omer
  Levy, Mike Lewis, Luke Zettlemoyer, and Veselin Stoyanov. 2019{\natexlab{b}}.
\newblock Roberta: A robustly optimized bert pretraining approach.
\newblock \emph{arXiv preprint arXiv:1907.11692}.

\bibitem[{Miyake and Shah(1999)}]{miyake1999models}
Akira Miyake and Priti Shah. 1999.
\newblock \emph{Models of working memory: Mechanisms of active maintenance and
  executive control}.
\newblock Cambridge University Press.

\bibitem[{Mostafazadeh et~al.(2016)Mostafazadeh, Chambers, He, Parikh, Batra,
  Vanderwende, Kohli, and Allen}]{mostafazadeh2016corpus}
Nasrin Mostafazadeh, Nathanael Chambers, Xiaodong He, Devi Parikh, Dhruv Batra,
  Lucy Vanderwende, Pushmeet Kohli, and James Allen. 2016.
\newblock A corpus and evaluation framework for deeper understanding of
  commonsense stories.
\newblock \emph{arXiv preprint arXiv:1604.01696}.

\bibitem[{Ott et~al.(2019)Ott, Edunov, Baevski, Fan, Gross, Ng, Grangier, and
  Auli}]{ott2019fairseq}
Myle Ott, Sergey Edunov, Alexei Baevski, Angela Fan, Sam Gross, Nathan Ng,
  David Grangier, and Michael Auli. 2019.
\newblock fairseq: A fast, extensible toolkit for sequence modeling.
\newblock In \emph{Proceedings of NAACL-HLT 2019: Demonstrations}.

\bibitem[{Papineni et~al.(2002)Papineni, Roukos, Ward, and
  Zhu}]{papineni2002bleu}
Kishore Papineni, Salim Roukos, Todd Ward, and Wei-Jing Zhu. 2002.
\newblock Bleu: a method for automatic evaluation of machine translation.
\newblock In \emph{Proceedings of the 40th annual meeting of the Association
  for Computational Linguistics}, pages 311--318.

\bibitem[{Peters et~al.(2018)Peters, Neumann, Iyyer, Gardner, Clark, Lee, and
  Zettlemoyer}]{PetersNIGCLZ18}
Matthew~E. Peters, Mark Neumann, Mohit Iyyer, Matt Gardner, Christopher Clark,
  Kenton Lee, and Luke Zettlemoyer. 2018.
\newblock \href {https://doi.org/10.18653/v1/n18-1202} {Deep contextualized
  word representations}.
\newblock In \emph{Proceedings of the 2018 Conference of the North American
  Chapter of the Association for Computational Linguistics: Human Language
  Technologies, {NAACL-HLT} 2018, New Orleans, Louisiana, USA, June 1-6, 2018,
  Volume 1 (Long Papers)}, pages 2227--2237. Association for Computational
  Linguistics.

\bibitem[{Phang et~al.(2020)Phang, Yeres, Swanson, Liu, Tenney, Htut, Vania,
  Wang, and Bowman}]{phang2020jiant}
Jason Phang, Phil Yeres, Jesse Swanson, Haokun Liu, Ian~F. Tenney, Phu~Mon
  Htut, Clara Vania, Alex Wang, and Samuel~R. Bowman. 2020.
\newblock \texttt{jiant} 2.0: A software toolkit for research on
  general-purpose text understanding models.
\newblock \url{http://jiant.info/}.

\bibitem[{Qi et~al.(2020)Qi, Yan, Gong, Liu, Duan, Chen, Zhang, and
  Zhou}]{QiYGLDCZ020}
Weizhen Qi, Yu~Yan, Yeyun Gong, Dayiheng Liu, Nan Duan, Jiusheng Chen, Ruofei
  Zhang, and Ming Zhou. 2020.
\newblock \href {https://doi.org/10.18653/v1/2020.findings-emnlp.217}
  {Prophetnet: Predicting future n-gram for sequence-to-sequence pre-training}.
\newblock In \emph{Proceedings of the 2020 Conference on Empirical Methods in
  Natural Language Processing: Findings, {EMNLP} 2020, Online Event, 16-20
  November 2020}, pages 2401--2410. Association for Computational Linguistics.

\bibitem[{Radford et~al.(2018)Radford, Narasimhan, Salimans, and
  Sutskever}]{radford2018improving}
Alec Radford, Karthik Narasimhan, Tim Salimans, and Ilya Sutskever. 2018.
\newblock Improving language understanding by generative pre-training.

\bibitem[{Radford et~al.(2019)Radford, Wu, Child, Luan, Amodei, and
  Sutskever}]{radford2019language}
Alec Radford, Jeffrey Wu, Rewon Child, David Luan, Dario Amodei, and Ilya
  Sutskever. 2019.
\newblock Language models are unsupervised multitask learners.
\newblock \emph{OpenAI blog}, 1(8):9.

\bibitem[{Raffel et~al.(2020)Raffel, Shazeer, Roberts, Lee, Narang, Matena,
  Zhou, Li, and Liu}]{RaffelSRLNMZLL20}
Colin Raffel, Noam Shazeer, Adam Roberts, Katherine Lee, Sharan Narang, Michael
  Matena, Yanqi Zhou, Wei Li, and Peter~J. Liu. 2020.
\newblock \href {http://jmlr.org/papers/v21/20-074.html} {Exploring the limits
  of transfer learning with a unified text-to-text transformer}.
\newblock \emph{J. Mach. Learn. Res.}, 21:140:1--140:67.

\bibitem[{Rajpurkar et~al.(2018)Rajpurkar, Jia, and Liang}]{RajpurkarJL18}
Pranav Rajpurkar, Robin Jia, and Percy Liang. 2018.
\newblock \href {https://www.aclweb.org/anthology/P18-2124/} {Know what you
  don't know: Unanswerable questions for squad}.
\newblock In \emph{Proceedings of the 56th Annual Meeting of the Association
  for Computational Linguistics, {ACL} 2018, Melbourne, Australia, July 15-20,
  2018, Volume 2: Short Papers}, pages 784--789. Association for Computational
  Linguistics.

\bibitem[{Rajpurkar et~al.(2016)Rajpurkar, Zhang, Lopyrev, and
  Liang}]{RajpurkarZLL16}
Pranav Rajpurkar, Jian Zhang, Konstantin Lopyrev, and Percy Liang. 2016.
\newblock \href {https://doi.org/10.18653/v1/d16-1264} {Squad: 100, 000+
  questions for machine comprehension of text}.
\newblock In \emph{Proceedings of the 2016 Conference on Empirical Methods in
  Natural Language Processing, {EMNLP} 2016, Austin, Texas, USA, November 1-4,
  2016}, pages 2383--2392. The Association for Computational Linguistics.

\bibitem[{Rush et~al.(2015)Rush, Chopra, and Weston}]{RushCW15}
Alexander~M. Rush, Sumit Chopra, and Jason Weston. 2015.
\newblock \href {https://doi.org/10.18653/v1/d15-1044} {A neural attention
  model for abstractive sentence summarization}.
\newblock In \emph{Proceedings of the 2015 Conference on Empirical Methods in
  Natural Language Processing, {EMNLP} 2015, Lisbon, Portugal, September 17-21,
  2015}, pages 379--389. The Association for Computational Linguistics.

\bibitem[{Sap et~al.(2020)Sap, Shwartz, Bosselut, Choi, and Roth}]{SapSBCR20}
Maarten Sap, Vered Shwartz, Antoine Bosselut, Yejin Choi, and Dan Roth. 2020.
\newblock \href {https://doi.org/10.18653/v1/2020.acl-tutorials.7} {Commonsense
  reasoning for natural language processing}.
\newblock In \emph{Proceedings of the 58th Annual Meeting of the Association
  for Computational Linguistics: Tutorial Abstracts, {ACL} 2020, Online, July
  5, 2020}, pages 27--33. Association for Computational Linguistics.

\bibitem[{Talmor et~al.(2020)Talmor, Elazar, Goldberg, and
  Berant}]{TalmorEGB20}
Alon Talmor, Yanai Elazar, Yoav Goldberg, and Jonathan Berant. 2020.
\newblock \href {https://transacl.org/ojs/index.php/tacl/article/view/2041}
  {olmpics - on what language model pre-training captures}.
\newblock \emph{Trans. Assoc. Comput. Linguistics}, 8:743--758.

\bibitem[{Talmor et~al.(2019)Talmor, Herzig, Lourie, and Berant}]{TalmorHLB19}
Alon Talmor, Jonathan Herzig, Nicholas Lourie, and Jonathan Berant. 2019.
\newblock \href {https://doi.org/10.18653/v1/n19-1421} {Commonsenseqa: {A}
  question answering challenge targeting commonsense knowledge}.
\newblock In \emph{Proceedings of the 2019 Conference of the North American
  Chapter of the Association for Computational Linguistics: Human Language
  Technologies, {NAACL-HLT} 2019, Minneapolis, MN, USA, June 2-7, 2019, Volume
  1 (Long and Short Papers)}, pages 4149--4158. Association for Computational
  Linguistics.

\bibitem[{Tenney et~al.(2019)Tenney, Xia, Chen, Wang, Poliak, McCoy, Kim,
  Van~Durme, Bowman, Das et~al.}]{tenney2019you}
Ian Tenney, Patrick Xia, Berlin Chen, Alex Wang, Adam Poliak, R~Thomas McCoy,
  Najoung Kim, Benjamin Van~Durme, Samuel~R Bowman, Dipanjan Das, et~al. 2019.
\newblock What do you learn from context? probing for sentence structure in
  contextualized word representations.
\newblock \emph{arXiv preprint arXiv:1905.06316}.

\bibitem[{Turing(2009)}]{turing2009computing}
Alan~M Turing. 2009.
\newblock Computing machinery and intelligence.
\newblock In \emph{Parsing the turing test}, pages 23--65. Springer.

\bibitem[{Vaswani et~al.(2017)Vaswani, Shazeer, Parmar, Uszkoreit, Jones,
  Gomez, Kaiser, and Polosukhin}]{VaswaniSPUJGKP17}
Ashish Vaswani, Noam Shazeer, Niki Parmar, Jakob Uszkoreit, Llion Jones,
  Aidan~N. Gomez, Lukasz Kaiser, and Illia Polosukhin. 2017.
\newblock \href
  {https://proceedings.neurips.cc/paper/2017/hash/3f5ee243547dee91fbd053c1c4a845aa-Abstract.html}
  {Attention is all you need}.
\newblock In \emph{Advances in Neural Information Processing Systems 30: Annual
  Conference on Neural Information Processing Systems 2017, December 4-9, 2017,
  Long Beach, CA, {USA}}, pages 5998--6008.

\bibitem[{Walton(2014)}]{walton2014abductive}
Douglas Walton. 2014.
\newblock \emph{Abductive reasoning}.
\newblock University of Alabama Press.

\bibitem[{Wang et~al.(2019{\natexlab{a}})Wang, Pruksachatkun, Nangia, Singh,
  Michael, Hill, Levy, and Bowman}]{WangPNSMHLB19}
Alex Wang, Yada Pruksachatkun, Nikita Nangia, Amanpreet Singh, Julian Michael,
  Felix Hill, Omer Levy, and Samuel~R. Bowman. 2019{\natexlab{a}}.
\newblock \href
  {https://proceedings.neurips.cc/paper/2019/hash/4496bf24afe7fab6f046bf4923da8de6-Abstract.html}
  {Superglue: {A} stickier benchmark for general-purpose language understanding
  systems}.
\newblock In \emph{Advances in Neural Information Processing Systems 32: Annual
  Conference on Neural Information Processing Systems 2019, NeurIPS 2019,
  December 8-14, 2019, Vancouver, BC, Canada}, pages 3261--3275.

\bibitem[{Wang et~al.(2019{\natexlab{b}})Wang, Singh, Michael, Hill, Levy, and
  Bowman}]{WangSMHLB19}
Alex Wang, Amanpreet Singh, Julian Michael, Felix Hill, Omer Levy, and
  Samuel~R. Bowman. 2019{\natexlab{b}}.
\newblock \href {https://openreview.net/forum?id=rJ4km2R5t7} {{GLUE:} {A}
  multi-task benchmark and analysis platform for natural language
  understanding}.
\newblock In \emph{7th International Conference on Learning Representations,
  {ICLR} 2019, New Orleans, LA, USA, May 6-9, 2019}. OpenReview.net.

\bibitem[{Wang et~al.(2019{\natexlab{c}})Wang, Liang, Zhang, Li, and
  Gao}]{WangLZLG19}
Cunxiang Wang, Shuailong Liang, Yue Zhang, Xiaonan Li, and Tian Gao.
  2019{\natexlab{c}}.
\newblock \href {https://doi.org/10.18653/v1/p19-1393} {Does it make sense? and
  why? {A} pilot study for sense making and explanation}.
\newblock In \emph{Proceedings of the 57th Conference of the Association for
  Computational Linguistics, {ACL} 2019, Florence, Italy, July 28- August 2,
  2019, Volume 1: Long Papers}, pages 4020--4026. Association for Computational
  Linguistics.

\bibitem[{Wang et~al.(2021)Wang, Fang, Khabsa, Mao, and
  Ma}]{wang2021entailment}
Sinong Wang, Han Fang, Madian Khabsa, Hanzi Mao, and Hao Ma. 2021.
\newblock Entailment as few-shot learner.
\newblock \emph{arXiv preprint arXiv:2104.14690}.

\bibitem[{Wolf et~al.(2020)Wolf, Debut, Sanh, Chaumond, Delangue, Moi, Cistac,
  Rault, Louf, Funtowicz, Davison, Shleifer, von Platen, Ma, Jernite, Plu, Xu,
  Scao, Gugger, Drame, Lhoest, and Rush}]{wolf-etal-2020-transformers}
Thomas Wolf, Lysandre Debut, Victor Sanh, Julien Chaumond, Clement Delangue,
  Anthony Moi, Pierric Cistac, Tim Rault, Rémi Louf, Morgan Funtowicz, Joe
  Davison, Sam Shleifer, Patrick von Platen, Clara Ma, Yacine Jernite, Julien
  Plu, Canwen Xu, Teven~Le Scao, Sylvain Gugger, Mariama Drame, Quentin Lhoest,
  and Alexander~M. Rush. 2020.
\newblock \href {https://www.aclweb.org/anthology/2020.emnlp-demos.6}
  {Transformers: State-of-the-art natural language processing}.
\newblock In \emph{Proceedings of the 2020 Conference on Empirical Methods in
  Natural Language Processing: System Demonstrations}, pages 38--45, Online.
  Association for Computational Linguistics.

\bibitem[{Yang et~al.(2019)Yang, Dai, Yang, Carbonell, Salakhutdinov, and
  Le}]{YangDYCSL19}
Zhilin Yang, Zihang Dai, Yiming Yang, Jaime~G. Carbonell, Ruslan Salakhutdinov,
  and Quoc~V. Le. 2019.
\newblock \href
  {https://proceedings.neurips.cc/paper/2019/hash/dc6a7e655d7e5840e66733e9ee67cc69-Abstract.html}
  {Xlnet: Generalized autoregressive pretraining for language understanding}.
\newblock In \emph{Advances in Neural Information Processing Systems 32: Annual
  Conference on Neural Information Processing Systems 2019, NeurIPS 2019,
  December 8-14, 2019, Vancouver, BC, Canada}, pages 5754--5764.

\bibitem[{Zellers et~al.(2018)Zellers, Bisk, Schwartz, and Choi}]{ZellersBSC18}
Rowan Zellers, Yonatan Bisk, Roy Schwartz, and Yejin Choi. 2018.
\newblock \href {https://doi.org/10.18653/v1/d18-1009} {{SWAG:} {A} large-scale
  adversarial dataset for grounded commonsense inference}.
\newblock In \emph{Proceedings of the 2018 Conference on Empirical Methods in
  Natural Language Processing, Brussels, Belgium, October 31 - November 4,
  2018}, pages 93--104. Association for Computational Linguistics.

\bibitem[{Zellers et~al.(2019)Zellers, Holtzman, Bisk, Farhadi, and
  Choi}]{ZellersHBFC19}
Rowan Zellers, Ari Holtzman, Yonatan Bisk, Ali Farhadi, and Yejin Choi. 2019.
\newblock \href {https://doi.org/10.18653/v1/p19-1472} {Hellaswag: Can a
  machine really finish your sentence?}
\newblock In \emph{Proceedings of the 57th Conference of the Association for
  Computational Linguistics, {ACL} 2019, Florence, Italy, July 28- August 2,
  2019, Volume 1: Long Papers}, pages 4791--4800. Association for Computational
  Linguistics.

\bibitem[{Zhang et~al.(2019)Zhang, Han, Liu, Jiang, Sun, and
  Liu}]{ZhangHLJSL19}
Zhengyan Zhang, Xu~Han, Zhiyuan Liu, Xin Jiang, Maosong Sun, and Qun Liu. 2019.
\newblock \href {https://doi.org/10.18653/v1/p19-1139} {{ERNIE:} enhanced
  language representation with informative entities}.
\newblock In \emph{Proceedings of the 57th Conference of the Association for
  Computational Linguistics, {ACL} 2019, Florence, Italy, July 28- August 2,
  2019, Volume 1: Long Papers}, pages 1441--1451. Association for Computational
  Linguistics.

\bibitem[{Zhou et~al.(2020)Zhou, Zhang, Cui, and Huang}]{ZhouZCH20}
Xuhui Zhou, Yue Zhang, Leyang Cui, and Dandan Huang. 2020.
\newblock \href {https://aaai.org/ojs/index.php/AAAI/article/view/6523}
  {Evaluating commonsense in pre-trained language models}.
\newblock In \emph{The Thirty-Fourth {AAAI} Conference on Artificial
  Intelligence, {AAAI} 2020, The Thirty-Second Innovative Applications of
  Artificial Intelligence Conference, {IAAI} 2020, The Tenth {AAAI} Symposium
  on Educational Advances in Artificial Intelligence, {EAAI} 2020, New York,
  NY, USA, February 7-12, 2020}, pages 9733--9740. {AAAI} Press.

\bibitem[{Zou et~al.(2021)Zou, Yin, Zhong, Yang, Yang, and
  Tang}]{zou2021controllable}
Xu~Zou, Da~Yin, Qingyang Zhong, Hongxia Yang, Zhilin Yang, and Jie Tang. 2021.
\newblock Controllable generation from pre-trained language models via inverse
  prompting.
\newblock \emph{arXiv preprint arXiv:2103.10685}.

\end{thebibliography}
\bibliographystyle{acl_natbib}

\newpage
%
\twocolumn[
\begin{@twocolumnfalse}
\section*{\Large{\centering{Supplementary Material for \emph{ElitePLM\\[25pt]}}}}
\end{@twocolumnfalse}
]

%
%

%
\appendix

We give some experiment-related information as supplementary materials. The appendix is organized into six sections:
\begin{itemize}
	\item Configurations and pretraining setting comparisons for selected models are presented in Appendix~\ref{configuration};
	\item Data statistics of each test are presented in Appendix~\ref{statistics};
	\item Full results for memory tests are presented in Appendix~\ref{memory};
	\item Full results for comprehension tests are presented in Appendix~\ref{comprehension};
	\item Full results for reasoning tests are presented in Appendix~\ref{reasoning}; and
	\item Full results for composition tests are presented in Appendix~\ref{composition}. 
\end{itemize}


\section{Configurations of Pretrained Language Models}
\label{configuration}

The selected ten PLMs within five categories and the comparisons of these PLMs in configuration and pretraining setting have been shown in Table~\ref{tab:models}. The effect extent of each factor for PLMs abilities in Table~\ref{tab:score}.

\section{Data Statistics}
\label{statistics}

\paratitle{Memory Tests.} The data statistics of LAMA and Wikipedia of each model are presented in Table~\ref{tab:memory-data}. Due to the differences of each PLM, we drop the data that are not in the vocabulary.

\paratitle{Comprehension Tests.} The data statistics of GLUE, SuperGLUE, SQuAD and RACE are presented in Table~\ref{tab:comprehension-data}.

\paratitle{Reasoning Tests.} The data statistics for commonsense reasoning, deductive reasoning and abductive reasoning are presented in Table~\ref{tab:reasoning-data}.

\paratitle{Composition Tests.} The data statistics for text summarization, question generation and story generation are presented in Table~\ref{tab:composition-data}. For the first three datasets, we truncate the source text considering the input length of PLMs during training. And for WritingPrompts, we reconstruct the original dataset and discard examples where text contains more than 512 tokens.

\begin{table*}[]
	\small
	\centering
	\begin{tabular}{c|c|cc|cc}
		\toprule
		\multirow{2.5}{*}{\textbf{Type}} & \multirow{2.5}{*}{\textbf{Models}} & \multicolumn{2}{c|}{\textbf{Configurations}} & \multicolumn{2}{c}{\textbf{Pretraining Setting}} \\ \cmidrule{3-6}
		&  & \textbf{Size}  & \textbf{\#Parameter} & \textbf{Corpus} & \textbf{Size} \\ 
		\midrule
		\multirow{4}{*}{Bidirectional} & BERT & base/large & 110M/340M & BooksCorpus, English Wikipedia & 16GB \\
		& RoBERTa & base/large & 125M/355M & \tabincell{c}{BooksCorpus, CC-News,\\ WebText, Stories} & 160GB \\
		& ALBERT & xlarge/xxlarge & 60M/235M & BERT Corpus & 16GB \\ 
		\midrule
		Unidirectional & GPT-2 & small/medium & 117M/345M & WebText (removing Wikipedia) & 40GB \\
		\midrule
		\multirow{2}{*}{Hybrid} & XLNet & base/large & 110M/340M & \tabincell{c}{BooksCorpus, English Wikipedia,\\ Giga5, ClueWeb, Common Crawl} & 158GB \\
		& UniLM & base/large & 110M/340M & BERT Corpus & 16GB \\
		\midrule
		\tabincell{c}{Knowledge-\\Enhanced} & ERNIE & base & 114M & English Wikipedia, Wikipedia & 17GB \\
		\midrule
		\multirow{3}{*}{Text-to-Text} & T5 & base/large & 220M/770M & Colossal Clean Crawled Corpus & 745GB \\
		& BART & base/large & 140M/400M & RoBERTa Corpus & 160GB \\
		& ProphetNet & large & 373M & RoBERTa Corpus & 160GB \\
		\bottomrule   
	\end{tabular}
	\caption{Configurations and pretraining setting comparisons for our selected models.}
	\label{tab:models}
\end{table*}

\begin{table*}[t]
 \small
 \centering
 \begin{tabular}{c|c|c|c|c|c}
 \toprule
  \textbf{Ability} & \textbf{MA} & \textbf{DD} & \textbf{MS} & \textbf{PO} & \textbf{PS} \\ 
                         \midrule
  \textbf{Memory}        & $\bigstar\bigstar$    &  $\bigstar$  &  $\bigstar$ &        $\bigstar\bigstar\bigstar$ & $\bigstar\bigstar\bigstar$ \\ \midrule
  \textbf{Comprehension} &    $\bigstar\bigstar$    &  $\bigstar\bigstar$   & $\bigstar$  & $\bigstar\bigstar$ &    $\bigstar\bigstar\bigstar$     \\ \midrule
  \textbf{Reasoning}     &  $\bigstar$ &   $\bigstar\bigstar\bigstar$  & $\bigstar$  &  $\bigstar\bigstar\bigstar$  &  $\bigstar\bigstar\bigstar$   \\ \hline
  \textbf{Composition}   &  $\bigstar$  &   $\bigstar\bigstar\bigstar$  & $\bigstar\bigstar\bigstar$  &  $\bigstar\bigstar\bigstar$  &   $\bigstar$    \\ 
  \bottomrule
 \end{tabular}
 \caption{The impact extent of each factor for four language abilities of PLMs. MA, DD, MS, PO, and PS are short for model architecture, data distribution, model size, pretraining objective, and pretraining strategy, respectively}
 \label{tab:score}
\end{table*}

\begin{table*}[]
\centering
\begin{tabular}{lrrrrr}
\toprule
\textbf{}                     & \multicolumn{1}{c}{\textbf{G-RE}} & \multicolumn{1}{c}{\textbf{T-REx}} & \multicolumn{1}{c}{\textbf{ConceptNet}} & \multicolumn{1}{c}{\textbf{SQuAD}} & \multicolumn{1}{c}{\textbf{Wikipedia}} \\ \midrule
\textbf{\#Origin}             & 6,106                             & 34,014                             & 14,878                                  & 305                                & 100,000                                \\
\textbf{\#Relation}             & 3                             & 41                             & 16                                  & -                                & -                                \\
\cmidrule{1-6}
\textbf{BERT / UniLM} & 5,527                             & 34,014                             & 11,658                                  & 305                                & 85,836                                 \\
\textbf{RoBERTa}              & 4,618                             & 29,500                             & 12,505                                  & 286                                & 85,862                                 \\
\textbf{ALBERT}               & 5,469                             & 33,636                             & 12,389                                  & 291                                & 86,533                                 \\
\textbf{ERNIE}                & 1,900                             & 9,071                              & 11,649                                  & 173                                & -                                      \\
\textbf{BART}                 & 4,618                             & 29,500                             & 12,505                                  & 286                                & 85,862                                 \\
\textbf{T5}                   & 4,256                             & 25,850                             & 10,905                                  & 230                                & 78,069                                 \\
\textbf{GPT-2}                 & 4,618                             & 29,500                             & 7,477                                   & 196                                & 1,184                                  \\
\textbf{XLNet}                & 5,202                             & 32,293                             & 12,080                                  & 279                                & 85,228                                 \\
\textbf{ProphetNet}           & 5,527                             & 34,014                             & 12,506                                  & 305                                & 87,516                                 \\ 
\cmidrule{1-6}
\textbf{The Predicted Outputs} & \multicolumn{5}{c}{The predicted token of ``\texttt{[MASK]}'' in each template.} \\
\bottomrule
\end{tabular}
\caption{Statistics of datasets in memory tests, including LAMA and Wikipedia. \#Origin and \#Relation denote the number of examples and relations in original dataset, and the number of each model denotes the number of examples after selected. The predicted outputs is the intermediate result resources we provide.}
	\label{tab:memory-data}
\end{table*}

\begin{table*}[]
\small
\centering	
\begin{tabular}{lrrrrl}
\toprule
\multicolumn{2}{c}{\textbf{Corpus}}                       & \multicolumn{1}{c}{\textbf{\#Train}} & \multicolumn{1}{c}{\textbf{\#Valid}} & \multicolumn{1}{c}{\textbf{\#Test}} & \multicolumn{1}{c}{\textbf{The Predicted Outputs}} \\ \midrule
\multirow{9}{*}{\textbf{GLUE}} 
& \textbf{CoLA}                         & 8,551                    & 1,043  & 1,063 & The predicted binary class whether a sentence is grammatical.   \\  \cline{2-6}
& \textbf{SST-2}                        & 67,349                   & 872    & 1,821 & The predicted sentiment (positive/negative) of a sentence.  \\ \cline{2-6}
& \textbf{MRPC}                         & 3,668                    & 408    & 1,725 & The predicted binary class whether two sentences are \\
& \textbf{QQP}                          & 363,846                  & 40,430 & 390,965 & semantically equivalent. \\ \cline{2-6}
& \textbf{STS-B}                        & 5,749                    & 1,500  & 1,379 & The predicted similarity score (1-5) of two sentences.  \\ \cline{2-6}
& \textbf{MNLI-M.}                 & \multirow{2}{*}{392,702} & 9,815  & 9,796   & The predicted relation (entailment/contradiction/neutral) \\
& \textbf{MNLI-MM.}              &                          & 9,832  & 9,847 & between two sentences.  \\ \cline{2-6}
& \textbf{QNLI}                         & 104,743                  & 5,463  & 5,463 &   \\
& \textbf{RTE}                          & 2,490                    & 277    & 3,000 & The predicted relation (entailment or not) between two sentences. \\
& \textbf{WNLI}                         & 635                      & 71     & 146 &  \\
\cmidrule{1-6}

\multirow{10}{*}{\textbf{SuperGLUE}} 
& \textbf{BoolQ}                        & 9,427                    & 3,270  & 3,245 & The predicted answer (yes/no) to the passage-based question.  \\ \cline{2-6}
& \multirow{2}{*}{\textbf{CB}}                           & \multirow{2}{*}{250}                      & \multirow{2}{*}{57}     & \multirow{2}{*}{250} & The predicted relation (entailment/contradiction/neutral) \\
& & & & & between two sentences.  \\\cline{2-6}
& \textbf{COPA}                         & 400                      & 100    & 500 & The predicted the cause or effect of the premise from two choices.    \\\cline{2-6}
& \textbf{MultiRC}                      & 5,100                    & 953    & 1,800 & The predicted answer choice to the passage-based question.  \\\cline{2-6}
& \multirow{2}{*}{\textbf{Wic}}                          & \multirow{2}{*}{6,000}                    & \multirow{2}{*}{638}    & \multirow{2}{*}{1,400} & The predicted binary class whether a word is used with the same \\
& & & & & sense in two sentences .  \\\cline{2-6}
& \textbf{WNLI}                         & 635                      & 71     & 146  & The predicted relation (entailment or not) between two sentences.   \\\cline{2-6}
& \multirow{2}{*}{\textbf{WSC}}                          & \multirow{2}{*}{554}                      & \multirow{2}{*}{104}    & \multirow{2}{*}{146}  & The predicted noun phrase referrent of the pronoun from among \\
& & & & & the provided choices.   \\
\cmidrule{1-6}
\multirow{2}{*}{\textbf{SQuAD}} & \textbf{v1.1}                   & 88,567                   & 10,790 & -  & \multirow{2}{*}{The predicted answer span to the passage-based question.}     \\
& \textbf{v2.0}                   & 131,924                  & 12,165 & -       \\
\cmidrule{1-6}
\multirow{6}{*}{\textbf{RACE}} & \multirow{2}{*}{\textbf{all}}    & 25,137                   & 1,389  & 1,407 & \multirow{6}{*}{The predicted answer choice to the passage-based question.}  \\
&                                       & 87,866                   & 4,887  & 4,934   \\
& \multirow{2}{*}{\textbf{middle}} & 6,409                    & 368    & 362     \\
&                                       & 25,421                   & 1,436  & 1,436   \\
& \multirow{2}{*}{\textbf{high}}   & 18,728                   & 1,021  & 1,045   \\
&                                       & 62,445                   & 3,451  & 3,498   \\
\bottomrule
\end{tabular}
\caption{Statistics of datasets in comprehension tests including GLUE, SuperGLUE, SQuAD and RACE. \#Train, \#Valid and \#Test denote the number of instances in train, valid and test set, respectively (the same as below). MNLI-M. and MNLI-MM. denote MNLI-match and MNLI-mismatch, respectively. SQuAD doesn't have test set, and we utilize the valid set as the test set. The predicted outputs is the intermediate result resources we provide.}
\label{tab:comprehension-data}
\end{table*}

\begin{table*}[]
\small
\centering
\begin{tabular}{llrrrl}
\toprule
\textbf{Reasoning Task}        & \textbf{Corpus}        & \multicolumn{1}{c}{\textbf{\#Train}} & \multicolumn{1}{c}{\textbf{\#Valid}} & \multicolumn{1}{c}{\textbf{\#Test}} & \multicolumn{1}{c}{\textbf{The Predicted Outputs}} \\ \midrule
\textbf{Commonsense} & \textbf{CQA} & 9,741                              & 1,221                              & 1,140  & The predicted answer choice to a commonsense question.                            \\
\cmidrule{1-6}
\multirow{4}{*}{\textbf{Deductive}} & \textbf{ROCStories}    & 1,257                              & 314                                & 1,571    & The predicted ending choice based on the context.                         \\
& \textbf{SWAG}          & 73,546                             & 20,006                             & 20,005           & The predicted answer choice based the grounded situation.                 \\
& \textbf{HellaSwag}     & 39,905                             & 10,042                             & 10,003                            \\
& \textbf{SM-A}           & 10,000                             & 1,000                              & 1,000 & The predicted valid sentence between two sentences.                            \\
\cmidrule{1-6}
\multirow{2}{*}{\textbf{Abductive}} & \textbf{SM-B}           & 10,000                             & 1,000                              & 1,000       &   The predicted reason choice why the sentence is invalid.                    \\
& \textbf{ARCT}          & 1,210                              & 316                                & 444   & The predicted warrant choice that justifies the reason and claim.                           \\ \bottomrule
\end{tabular}
\caption{Statistics of datasets in reasoning tests, including commonsense reasoning, deductive reasoning and abductive reasoning. CQA is short for CommonsenseQA. SM-A and SM-B denote the Task A and Task B of Sense Making, respectively. The Predicted outputs is the intermediate result resources we provide.}
	\label{tab:reasoning-data}
\end{table*}

\begin{table*}[]
\small
\centering
\begin{tabular}{llrrrrrl}
\toprule
\textbf{Task}         & \textbf{Corpus}         & \multicolumn{1}{c}{\textbf{\#Train}} & \multicolumn{1}{c}{\textbf{\#Valid}} & \multicolumn{1}{c}{\textbf{\#Test}} & \multicolumn{1}{c}{\textbf{\#Input}} & \multicolumn{1}{c}{\textbf{\#Output}}  & \multicolumn{1}{c}{\textbf{The Predicted Outputs}} \\ \midrule
\multirow{3}{*}{\textbf{TS}} & \textbf{CNN/Daily Mail} & 287,113                            & 13,368                             & 11,490                            & 822.3 & 57.9 & The generated summary given a news. \\
& \multirow{2}{*}{\textbf{Gigaword}}       & \multirow{2}{*}{3,803,957}                          & \multirow{2}{*}{189,651}                            & \multirow{2}{*}{1,951}                             & \multirow{2}{*}{33.7} & \multirow{2}{*}{8.7} & The generated headline given a paragraph and \\
&&&&&&& corresponding Turing test and aspect scores.\\
\cmidrule{1-8}
\multirow{2}{*}{\textbf{QG}} & \multirow{2}{*}{\textbf{SQuAD}}          & \multirow{2}{*}{75,722}                             & \multirow{2}{*}{10,570}                             & \multirow{2}{*}{11,877}                           & \multirow{2}{*}{149.4} & \multirow{2}{*}{11.5} & The generated question given a passage and \\
&&&&&&& corresponding answer.\\
\cmidrule{1-8}
\multirow{2}{*}{\textbf{SG}} & \multirow{2}{*}{\textbf{WritingPrompts}} & \multirow{2}{*}{67,765}                             & \multirow{2}{*}{3,952}                              & \multirow{2}{*}{3,784}                           & \multirow{2}{*}{30.2} & \multirow{2}{*}{281.2} & The generated story given a prompt and \\
&&&&&&& corresponding Turing test and aspect scores. \\ \bottomrule
\end{tabular}
\caption{Statistics of datasets in composition tests, including text summarization (TG), question generation (QG) and story generation (SG). \#Input and \#Output denote the average number of tokens in the input text and output text. The Predicted outputs is the intermediate results and human evaluation resources we provide.}
	\label{tab:composition-data}
\end{table*}

\section{Memory Tests}
\label{memory}

Full results on LAMA and Wikipedia datasets are presented in Table~\ref{tab:memory-app}.

\begin{table*}[h]
	\small
	\centering
	\begin{tabular}{lrrrrrrr}
		\toprule
		\textbf{Models} & \textbf{Vocab Size} & \textbf{LAMA-G} & \textbf{LAMA-T} & \textbf{LAMA-C} & \textbf{LAMA-S} & \textbf{Wikipedia} & \textbf{Average} \\
		\midrule
		BERT\textsubscript{\textsc{base}} & 28996               & {\ul 10.3}              & 27.5                & 15.3                     & 12.8                & 66.8               & 41.6             \\
		BERT\textsubscript{\textsc{large}} & 28996               & \textbf{11.0}           & \textbf{29.2}       & 19.1                     & 17.0                & 70.9               & {\ul 45.0}       \\
		RoBERTa\textsubscript{\textsc{base}} & 50265               & 7.5                     & 19.9                & 17.9                     & 13.3                & 66.9               & 40.8             \\
		RoBERTa\textsubscript{\textsc{large}} & 50265               & 7.1                     & 23.9                & \textbf{21.6}            & \textbf{21.0}       & {\ul 71.1}         & 44.8             \\
		ALBERT\textsubscript{\textsc{xlarge}} & 30000               & 2.9                     & 19.6                & 16.8                     & 14.4                & 64.3               & 38.9             \\
		ALBERT\textsubscript{\textsc{xxlarge}} & 30000               & 3.3                     & 21.0                & 20.0                     & {\ul 20.6}          & 63.9               & 40.1             \\
		GPT-2\textsubscript{\textsc{small}} & 50257               & 1.3                     & 6.8                 & 4.0                      & 3.0                 & 36.0               & 19.9             \\
		GPT-2\textsubscript{\textsc{medium}} & 50257               & 3.9                     & 12.0                & 6.4                      & 5.6                 & 42.7               & 24.8             \\
		XLNet\textsubscript{\textsc{base}} & 32000               & 0.0                     & 0.0                 & 2.8                      & 0.0                 & 64.6               & 32.7             \\
		XLNet\textsubscript{\textsc{large}} & 32000               & 0.0                     & 0.0                 & 5.5                      & 0.4                 & 68.7               & 35.1             \\
		UniLM\textsubscript{\textsc{base}} & 28996               & 8.5                     & 27.6                & 15.4                     & 11.8                & 66.9               & 41.4             \\
		UniLM\textsubscript{\textsc{large}} & 28996               & 9.6                     & {\ul 28.4}          & 18.3                     & 17.4                & \textbf{71.5}      & \textbf{46.4}    \\
		ERNIE\textsubscript{\textsc{base}} & 28996               & 1.3                     & 13.4                & 13.0                     & 8.1                 & -                  & -                \\
		T5\textsubscript{\textsc{base}} & 32100               & 5.5                     & 20.0                & 13.2                     & 9.6                 & 60.5               & 36.3             \\
		T5\textsubscript{\textsc{large}} & 32100               & 4.0                     & 21.7                & 17.1                     & 11.7                & 65.0               & 39.3             \\
		BART\textsubscript{\textsc{base}} & 50295               & 5.7                     & 11.7                & 9.5                      & 4.2                 & 47.9               & 27.8             \\
		BART\textsubscript{\textsc{large}} & 50295               & 9.4                     & 15.8                & 7.7                      & 3.1                 & 47.8               & 28.4             \\
		ProphetNet\textsubscript{\textsc{large}} & 30522               & 0.1                     & 1.1                 & 0.3                      & 0.7                 & 31.3               & 15.9             \\
		\bottomrule
	\end{tabular}
	\caption{Memory tests results on LAMA and Wikipedia datasets (test set). We report accuracy score for each dataset. Average is computed by averaging the scores of LAMA and Wikipedia (the score of LAMA is averaged among four dataset first). LAMA-G, LAMA-T, LAMA-C and LAMA-S denote the LAMA corpus Google-RE, T-REx, ConceptNet and SQuAD, respectively.}
	\label{tab:memory-app}
\end{table*}

\section{Comprehension Tests}
\label{comprehension}

Full results on SuperGLUE, SQuAD and RACE are presented in Table~\ref{tab:superglue} and Table~\ref{tab:squadrace}.

\begin{table*}[]
	\centering
	\begin{tabular}{lcccccccc}
		\toprule
		\textbf{Model}   & \textbf{WSC}  & \textbf{CB}        & \textbf{RTE}  & \textbf{COPA} & \textbf{Wic}  & \textbf{BoolQ} & \textbf{MultiRC} & \textbf{Avg} \\
		& Acc.          & F1/Acc.            & Acc.          & Acc.          & Acc.          & Acc.           & F1/EM           &             \\
		\midrule
		BERT\textsubscript{\textsc{base}}        & 60.6          & 78.7/80.4          & 66.4          & 65.0          & 69.9          & 74.6           & 68.1/16.9       & 65.5       \\
		BERT\textsubscript{\textsc{large}}       & 63.5          & 89.0/92.9          & 70.1          & 73.0          & \underline{72.7}    & 75.6           & 69.4/22.6             & 70.3       \\
		RoBERTa\textsubscript{\textsc{base}}     & 71.1          & 89.1/91.1          & 75.1          & 78.0          & 67.2          & 81.1           & 72.6/31.9               & 73.6       \\
		RoBERTa\textsubscript{\textsc{large}}    & 75.0          & 95.0/96.4          & \underline{88.2}    & 84.0          & \underline{72.7}    & \textbf{85.4}  & 81.7/47.2               & \underline{80.8}             \\
		ALBERT\textsubscript{\textsc{xlarge}}    & 63.5          & 81.1/85.7          & 62.5          & 75.0          & 66.5          & 62.2           & 63.6/12.4  & 64.4   \\
		ALBERT\textsubscript{\textsc{xxlarge}}   & 64.4          & 87.6/92.9          & 70.4          & \textbf{91.0} & \textbf{74.3} & 62.2           & \textbf{85.1}/\textbf{54.0}  & 74.6   \\
		GPT-2\textsubscript{\textsc{small}}        & 54.8          & 64.0/76.8          & 62.1          & 62.0          & 64.1          & 68.2           & 67.3/19.5   & 60.7   \\
		GPT-2\textsubscript{\textsc{medium}}       & 61.5          & 84.4/82.1          & 63.6          & 63.0          & 67.2          & 73.9           & 71.5/29.2   & 66.1   \\
		XLNet\textsubscript{\textsc{base}}       & 64.4          & 91.0/91.1          & 59.9          & 65.0          & 67.9          & 76.9           & 72.5/29.6  & 68.0   \\
		XLNet\textsubscript{\textsc{large}}      & 65.3          & 87.6/92.9          & \textbf{88.5} & 82.0          & 69.7          & 84.7           & 79.0/41.6   & 77.3   \\
		UniLM\textsubscript{\textsc{base}}       & 63.5          & 74.7/82.1          & 60.3          & 67.0          & 68.5          & 73.3           & 67.9/20.5  & 65.0   \\
		UniLM\textsubscript{\textsc{large}}      & 65.4          & 86.5/87.5          & 70.9          & 76.0          & 72.3          & 82.3           & 75.7/36.3   & 72.8   \\
		ERNIE\textsubscript{\textsc{base}}       & 65.4          & 81.6/82.1          & 68.8          & 64.0          & 70.8          & 74.4           & 68.7/21.3  & 67.2   \\
		T5\textsubscript{\textsc{base}}          & \underline{79.8}    & 86.2/94.0          & 80.1          & 71.2          & 68.3          & 81.4           & 79.7/43.1             & 76.0       \\
		T5\textsubscript{\textsc{large}}         & \textbf{84.6} & 91.6/94.8          & 87.2          & 83.4          & 69.3          & \textbf{85.4}  & \underline{83.3}/\underline{50.7}              & \textbf{81.4}                    \\
		BART\textsubscript{\textsc{base}}        & 64.4          & 86.6/85.7          & 69.5          & 70.0          & 65.7          & 75.7           & 74.2/31.7  & 69.2   \\
		BART\textsubscript{\textsc{large}}       & 65.4          & \textbf{97.4}/\textbf{96.4} & 83.5          & \underline{86.0}    & 70.4          & 85.1              &      82.9/50.6           & 79.2   \\
		ProphetNet\textsubscript{\textsc{large}} & 63.5          & \underline{94.7}/\underline{92.9}    & 51.3          & 61.0          & 60.7          & 67.4             &      64.7/17.2            & 62.7        \\
		\bottomrule                
	\end{tabular}
	\caption{Comprehension tests results on SuperGLUE (valid set). Avg column is computed by averaging the scores of tasks to its left (the scores for CB and MultiRC are first averaged).}
	\label{tab:superglue}
\end{table*}

\begin{table*}[]
	\centering
	\begin{tabular}{lccccccc}
		\toprule
		\multirow{2.5}{*}{\textbf{Models}} & \multicolumn{2}{c}{\textbf{SQuAD v1.1}} & \multicolumn{2}{c}{\textbf{SQuAD v2.0}} & \multicolumn{3}{c}{\textbf{RACE}}                   \\
		\cmidrule(lr){2-3}\cmidrule(lr){4-5}\cmidrule(lr){6-8}
		\multicolumn{1}{c}{}                                 & EM                 & F1                 & EM                 & F1                 & RACE          & RACE-M        & RACE-H              \\
		\midrule
		BERT\textsubscript{\textsc{base}}          & 80.8               & 88.5               & 72.8               & 76.0               & 65.0          & 71.7          & 62.3                \\
		BERT\textsubscript{\textsc{large}}         & 84.1               & 90.9               & 78.7               & 81.9               & 72.0          & 76.6          & 70.1                \\
		RoBERTa\textsubscript{\textsc{base}}       & 86.1               & 92.3               & 80.3               & 83.4               & 72.8          & 72.6          & 26.6                \\
		RoBERTa\textsubscript{\textsc{large}}      & {\ul 88.9}         & {\ul 94.6}         & {\ul 86.5}         & {\ul 89.4}         & 83.2          & {\ul 86.5}    & 81.3                \\
		ALBERT\textsubscript{\textsc{xlarge}}      & 86.1               & 92.5               & 83.1               & 86.1               & 78.1          & 76.7          & 79.8                \\
		ALBERT\textsubscript{\textsc{xxlarge}}     & 88.3               & 94.1               & 85.1               & 88.1               & \textbf{87.4} & 85.9          & \textbf{87.1}       \\
		GPT-2\textsubscript{\textsc{small}}          & 63.6               & 75.1               & 57.1               & 61.5               & 61.2          & 62.9          & 58.2                \\
		GPT-2\textsubscript{\textsc{medium}}         & 70.3               & 80.8               & 61.5               & 66.0               & 62.2          & 65.0          & 61.4                \\
		XLNet\textsubscript{\textsc{base}}         & 12.8               & 14.7               & 78.5               & 81.3               & 71.3          & 72.8          & 67.5                \\
		XLNet\textsubscript{\textsc{large}}        & \textbf{89.7}      & \textbf{95.1}      & \textbf{87.9}      & \textbf{90.6}      & {\ul 85.4}    & \textbf{88.6} & {\ul \textbf{84.0}} \\
		UniLM\textsubscript{\textsc{base}}         & 82.8               & 89.9               & 74.9               & 78.0               & 59.0          & 64.1          & 50.3                \\
		UniLM\textsubscript{\textsc{large}}        & 86.5               & 92.7               & 80.5               & 83.4               & 70.3          & 70.0          & 66.4                \\
		ERNIE\textsubscript{\textsc{base}}         & -                  & -                  & -                  & -                  & -             & 67.8          & -                   \\
		T5\textsubscript{\textsc{base}}            & 85.4               & 92.1               & 77.6               & 81.3               & 70.6          & 74.4          & 68.4                \\
		T5\textsubscript{\textsc{large}}           & 86.7               & 93.8               & -                  & -                  & 80.4          & 82.6          & 77.8                \\
		BART\textsubscript{\textsc{base}}          & 84.6               & 91.0               & 76.0               & 79.2               & 70.1          & 72.4          & 63.2                \\
		BART\textsubscript{\textsc{large}}         & 88.8               & {\ul 94.6}         & 86.1               & 89.2               & 82.2          & 82.5          & 79.6                \\
		ProphetNet\textsubscript{\textsc{large}}   &-                  & -                  & -                  & -                  & -       & 74.1          & -            \\
		\bottomrule
	\end{tabular}
	\caption{Comprehension tests results on SQuAD and RACE (test set).}
	\label{tab:squadrace}
\end{table*}

\section{Reasoning Tests}
\label{reasoning}

Full results on CommonsenseQA, ROCStories, SWAG, HellaSwag, Sense Making, and ARCT are presented in Table~\ref{tab:reasoning-app}.

\begin{table*}[t]
	\centering
	\begin{tabular}{lccccccc}
		\toprule
		\textbf{Model}   & \textbf{CQA} & \textbf{ROCStories} & \textbf{SWAG} & \textbf{HellaSwag} & \textbf{SM-A} & \textbf{SM-B} & \textbf{ARCT} \\
		\midrule
		BERT\textsubscript{\textsc{base}} & 53.0          & 88.1          & 81.6          & 40.5          & 87.3          & 80.1          & 65.1          \\
		BERT\textsubscript{\textsc{large}} & 55.9          & 90.2          & 86.3          & 47.3          & 89.4          & 85.8          & 71.2          \\
		RoBERTa\textsubscript{\textsc{base}} & 72.1          & 93.3          & 82.6          & 61.0          & 89.3          & 87.5          & 46.1          \\
		RoBERTa\textsubscript{\textsc{large}} & 72.2          & \textbf{97.4} & {\ul 89.9}    & {\ul 85.2}    & \textbf{93.0} & \textbf{92.3} & 57.9          \\
		ALBERT\textsubscript{\textsc{xlarge}} & 66.2          & 90.4          & 84.6          & 75.9          & 87.9          & 89.4          & 56.1          \\
		ALBERT\textsubscript{\textsc{xxlarge}} & \textbf{80.0} & {\ul 97.1}    & \textbf{90.7} & \textbf{90.1} & 92.5          & \textbf{92.3} & 79.5          \\
		GPT-2\textsubscript{\textsc{small}} & 47.8          & 58.8          & 48.1          & 39.9          & 84.2          & 74.7          & 66.0          \\
		GPT-2\textsubscript{\textsc{medium}} & 60.8          & 59.9          & 79.7          & 60.4          & 88.7          & 73.4          & 66.7          \\
		XLNet\textsubscript{\textsc{base}} & 53.8          & 92.0          & 80.4          & 55.1          & 81.6          & 85.4          & 80.2          \\
		XLNet\textsubscript{\textsc{large}} & 62.9          & 93.8          & 86.8          & 79.7          & 83.7          & 88.7          & {\ul 83.1}    \\
		UniLM\textsubscript{\textsc{base}} & 47.6          & 80.6          & 77.0          & 36.3          & 86.2          & 83.6          & 48.4          \\
		UniLM\textsubscript{\textsc{large}} & 62.3          & 86.9          & 83.1          & 46.7          & 89.3          & 86.4          & 72.3          \\
		ERNIE\textsubscript{\textsc{base}} & 54.1          & 84.7          & -             & -             & 88.7          & -             & 73.7          \\
		T5\textsubscript{\textsc{base}} & 61.9          & 88.2          & 65.8          & 55.2          & 89.2          & 82.9          & 63.3          \\
		T5\textsubscript{\textsc{large}} & 69.8          & 91.4          & 73.7          & 79.1          & {\ul 92.7}    & 88.2          & 69.4          \\
		BART\textsubscript{\textsc{base}} & 61.0          & 88.9          & 81.2          & 53.4          & 72.0          & 67.9          & 71.8          \\
		BART\textsubscript{\textsc{large}} & {\ul 75.8}    & 91.7          & 87.9          & 76.6          & 82.9          & 67.9          & \textbf{84.2} \\
		ProphetNet\textsubscript{\textsc{large}} & 21.3          & 82.2          & 70.1          & 26.4          & 85.5          & 78.0          & 65.5          \\
		\bottomrule
	\end{tabular}
	\caption{Reasoning tests results on seven datasets (test set). We report accuracy score for each dataset. CQA is short for CommonsenseQA. SM-A and SM-B denote the Task A and Task B of Sense Making, respectively.}
	\label{tab:reasoning-app}
\end{table*}

\section{Composition Tests}
\label{composition}

For automatic metrics, BLEU-$n$ and ROUGE-$n$ compute the ratios of overlapping $n$-grams between generated and real text, while METEOR measures word-to-word matches based on WordNet between generated and real text. For the human test, \textit{Fluency} evaluates whether the text is well-formed and logical to read; \textit{Informativeness} measures whether the text contains useful information; \textit{Accuracy} tests whether the text describes the given content accurately; \textit{Relevance} measures whether the text is relevant to the given context; \textit{Overall} evaluates the overall quality of the text. 

In the human test, we ramdomly select 500 generated texts for each PLM and 500 gold text. Therefore, there are 3000 texts totally. The judges are all PhD students which do not know about where each text comes from. Each text will be scored by two judges from the above five aspects, and the final score is the average of the two scores. In the Turing test, each text will also be distinguished by two judges. Only when two judges make the same decisions that the text is generated by human, we will consider the text is true.

Full results on CNN/Daily-Mail, GigaWord, SQuAD, and WritingPrompts are presented in Table~\ref{tab:composition-app}. Turing test results are presented in Table~\ref{tab:turing-test}. We also show some summaries and stories generated by different PLMs in Table~\ref{tab:gigaword-ex}, Table~\ref{tab:writingprompts-ex1}, and Table~\ref{tab:writingprompts-ex2}.

\begin{table*}[t]
	\small
	\centering
	\begin{tabular}{lrrrrrrrrrrrr}
		\toprule
		\multirow{2.5}*{\textbf{Models}} & \multicolumn{3}{c}{\textbf{CNN-DailyMail}} & \multicolumn{3}{c}{\textbf{GigaWord}} & \multicolumn{3}{c}{\textbf{SQuAD}} & \multicolumn{3}{c}{\textbf{WritingPrompts}} \\
		\cmidrule(r){2-4}\cmidrule(r){5-7}\cmidrule(r){8-10}\cmidrule(r){11-13}
		& \makecell[c]{R-1} & \makecell[c]{R-2} & \makecell[c]{R-L} & \makecell[c]{R-1} & \makecell[c]{R-2} & \makecell[c]{R-L} & \makecell[c]{B-4} & \makecell[c]{R-L} & \makecell[c]{ME} & \makecell[c]{B-4} & \makecell[c]{R-L} & \makecell[c]{ME} \\
		\midrule
		GPT-2\textsubscript{\textsc{small}}       &  24.60   &  7.21  &  21.06   & 25.25          & 9.03           & 23.20          & 5.13           & 14.83          & 21.06          & 11.58          & 3.80          & 8.18          \\
		GPT-2\textsubscript{\textsc{medium}}      &  22.95         &    5.99      &      22.08    & 23.72          & 8.12           & 21.56          & 8.48           & 18.82          & 26.77          & 14.47          & 3.23          & 7.29          \\
		UniLM\textsubscript{\textsc{base}}       & 17.83          & 0.11           & 5.50           &  16.64    &  6.11     &    15.12    & 4.47           & 17.65          & 20.30          & \textbf{27.71} & 2.35          & 5.47          \\
		UniLM\textsubscript{\textsc{large}}      & 43.44          & 20.21          & 40.51          & 38.45          & 19.45          & 35.75          & 4.42           & 17.43          & 20.13          & {\ul 26.88}    & 1.84          & 5.01          \\
		T5\textsubscript{\textsc{base}}          & 42.05          & 20.34          & 39.40          & 33.13          & 15.60          & 30.18          & 11.18          & 21.82          & 29.93          & 6.04           & \textbf{4.61} & \textbf{9.81} \\
		T5\textsubscript{\textsc{large}}         & 42.50          & 20.68          & 39.75          & 34.75          & 16.26          & 31.49          & 11.19          & 22.35          & 30.53          & 8.61           & {\ul 4.19}    & {\ul 9.51}    \\
		BART\textsubscript{\textsc{base}}        & 36.36          & 20.87          & 33.32          &  38.65     &  19.43  &   35.82  & {\ul 14.44}    & {\ul 24.11}    & {\ul 36.92}    & 11.91          & 3.57          & 7.69          \\
		BART\textsubscript{\textsc{large}}       & {\ul 44.16}    & \textbf{21.28} & {\ul 40.90}    & {\ul 39.41}    & {\ul 20.21}    & {\ul 36.42}    & \textbf{15.87} & \textbf{25.47} & \textbf{38.42} & 14.72          & 3.14          & 7.08          \\
		ProphetNet\textsubscript{\textsc{large}} & \textbf{44.20} & {\ul 21.17}    & \textbf{41.30} & \textbf{39.51} & \textbf{20.42} & \textbf{36.69} & 14.20          & 23.97          & 35.99          & 19.31          & 2.59          & 7.19         \\
		\bottomrule
	\end{tabular}
	\caption{Composition tests results on four datasets. R-1, R-2, R-L are short for ROUGE-1, ROUGE-2, ROUGE-L respectively. B-4 and MT denote BLEU-4 and METEOR, respectively.}
	\label{tab:composition-app}
\end{table*}

\begin{table*}[t]
	\centering
	\begin{tabular}{lr|rrrrr}
		\toprule
		 \textbf{Models} & \textbf{TT (\%)} & \textbf{Fluency} & \textbf{Informativeness} & \textbf{Accuracy} & \textbf{Coherence} & \textbf{Overall} \\
		\midrule
		GPT-2\textsubscript{\textsc{medium}} & \textbf{45.7} & \textbf{3.42} & \textbf{3.17} & {\ul 3.20}    & \textbf{3.23} & {\ul 5.87}    \\
		UniLM\textsubscript{\textsc{large}} & 1.2           & 1.32          & 1.88          & 2.03          & 1.71          & 2.74          \\
		T5\textsubscript{\textsc{large}} & 34.4          & 3.01          & 2.80          & 3.09          & 2.87          & 5.18          \\
		BART\textsubscript{\textsc{large}} & {\ul 45.2}    & 3.37          & {\ul 3.16}    & \textbf{3.39} & {\ul 3.22}    & \textbf{5.96} \\
		ProphetNet\textsubscript{\textsc{large}} & 29.6          & 2.95          & 2.91          & 3.10          & 2.89          & 5.18          \\
		\cmidrule{1-7}
		Gold   &    71.3          & 3.79          & 4.07          & 3.87          & 3.80          & 7.37           \\
		\bottomrule
	\end{tabular}
	\caption{Turing test and human scores on the test set of WritingPrompts. TT is short for Turing Test and represents the percentage of makers who believe the text is written by human. Coherence evaluates how content is coherent considering both intra- and inter-sentence correlation of a paragraph. Rated from 1 to 5.}
	\label{tab:turing-test-app}
\end{table*}

\begin{table*}[]
	\small
	\centering
	\begin{tabular}{lp{0.8\textwidth}}
		\toprule
		\textbf{Passage}    & japan 's nec corp. and UNK computer corp. of the united states said wednesday they had agreed to join forces in supercomputer sales .         \\ 
		\midrule                                                   
		\textbf{GPT-2\textsubscript{\textsc{medium}}}       & unk computer to jointly sell supercomputers                                                                                                                                                       \\
		\textbf{UniLM\textsubscript{\textsc{large}}}      & nec {[}UNK{]} to join forces in supercomputer sales                                                                                                                                               \\
		\textbf{T5\textsubscript{\textsc{large}}}         & nc and unk computer to join forces in supercomputer sales                                                                                                                                         \\
		\textbf{BART\textsubscript{\textsc{large}}}       & nec and unk computer to join forces in supercomputer sales                                                                                                                                        \\
		\textbf{ProphetNet\textsubscript{\textsc{large}}} & nec unk computer to join forces in supercomputer sales                                                                                                                                            \\
		\cmidrule{1-2}
		\textbf{Gold}       & nec UNK in computer sales tie-up                                                                                                                                                                  \\
		\midrule
		\textbf{Passage}    & the sri lankan government on wednesday announced the closure of government schools with immediate effect as a military campaign against tamil separatists escalated in the north of the country . \\
		\midrule
		\textbf{GPT-2\textsubscript{\textsc{medium}}}       & closed as military campaign escalates                                                                                                                                                             \\
		\textbf{UniLM\textsubscript{\textsc{large}}}      & sri lankan gov t shuts down schools in north                                                                                                                                                      \\
		\textbf{T5\textsubscript{\textsc{large}}}         & sri lanka closes schools as fighting escalates                                                                                                                                                    \\
		\textbf{BART\textsubscript{\textsc{large}}}       & sri lanka closes government schools                                                                                                                                                               \\
		\textbf{ProphetNet\textsubscript{\textsc{large}}} & sri lanka closes schools as military escalates                                                                                                                                                    \\
		\cmidrule{1-2}
		\textbf{Gold}       & sri lanka closes schools as war escalates                 \\             
		\bottomrule                                                                                                                          
	\end{tabular}
	\caption{Test set examples of GigaWord dataset for each model.}
	\label{tab:gigaword-ex}
\end{table*}

\begin{table*}[]
	\small
	\centering
	\begin{tabular}{lp{0.8\textwidth}}
		\toprule
		\textbf{Prompts}    & {[} WP {]} You have a very shitty type of precognition .                                                                                                                                                                                                                                                                                                                                                                                                                                                                                                                                                                                                                                                                                                                                                                                                                                                                                                                                                                                                                                                                                                                                                                                                                                                                                                                                                                                                                                                                                                                                                                                                                                                                                                                                                                                                                                                                                                                                                                                                                                                                                                                                                                                                                                                                                                                                                                                                                                                                                                                                                                                        \\ \midrule
		\textbf{GPT-2\textsubscript{\textsc{medium}}}       & \textless{}newline\textgreater `` I'm sorry, '' I said. `` I don't know what you're talking about. '' I looked around the room, trying to find the source of the noise. `` What do you mean? '' I asked. `` It's not like I've never heard it before. '' He looked at me, confused. `` No, '' he said, `` it's never happened to me before. I 'd be surprised if it ever did. '' `` What's that supposed to mean? You're telling me that I can't remember anything before? '' `` No. '' he replied. `` That's exactly what it means. You've always been precognitions, and you 'll never be able to remember anything. ''                                                                                                                                                                                                                                                                                                                                                                                                                                                                                                                                                                                                                                                                                                                                                                                                                                                                                                                                                                                                                                                                                                                                                                                                                                                                                                                                                                                                                                                                                                                                                                                                                                                                                                                                                                                                                                                                                                                                                                                                                       \\ 
		\cmidrule{1-2}
		\textbf{UniLM\textsubscript{\textsc{large}}}      & i'm not sure what i've done wrong. i'd rather not be able to do it. i don't know what i have done wrong, i don't know what it has done. i have been told that i have n've been wrong. \textless newline \textgreater \textless new york. \textless york \textgreater \textless yoline \textgreater i'll never forget that day. \textless \textless new \textgreater \textless \textless{}line \textgreater the day i was born. the day that i was raised. the one that i never forget. the time that i first met her. the moment that i met her, the time when i first saw her. \textless nowline \textgreater my life is a blur. the blur is the blur. i am not sure if i have ever been wrong, or if i had ever been right. i ca n're sure if it was the blur or the blur, but i do know it was. \textless tim. \textless the blur \textless new newline. \textless i am a blur, the blur and the blur are the same. the same, the same is the same thing. the difference is the difference between the two. the two are the difference. the differences are the differences. the different things. the things... the changes. the change. the changing. the thing.. is the change, the change is the thing that is the different. the new thing. \textless{}.. \textless this is the story of a man who was born in the blur of the two, the one where i was n's. the other where i am. the first time i met. the second time i saw her, and the first where i never forgot. the story is the one in the middle of the blur where i met the woman i loved. the woman who i loved, the woman that i loved the woman. the man who i love the woman with the most. the person who i hate the most, the person that i hate most. \textless cline \textgreater * * * \textless new story \textgreater \textless nbsp \textgreater i am the blur ; i am i. \textless{}pline \textgreater \textgreater \textless{}bsp. i was a blur ; \textless new line \textgreater \textless i was the one who was the first to be born. \textless first time. \textless np \textgreater \textless do nbs. \textless do ibs. i did n'i do it, i did not know it had happened. \textless{}line i woman i \textless i i i ve, the \textless blur ve                                                                                                                                                                                                                                                                                                                                                                                                                \\ \cmidrule{1-2}
		\textbf{T5\textsubscript{\textsc{large}}}         & \textless{}newline\textgreater It's been a long time since I've been able to use my powers. I haven't had a chance to use them in a while, but I do have a few things I can do to help. The first thing I do is go to the bathroom, and I have to wash my hands. It is a pain, I know it. But I do it anyways, because I have the ability to see things. And I can see them. They can see me in the mirror, they can see my eyes, their eyes. My hands, my feet, the way they move, how they feel. How I feel, what they do to me. What I do to them? How do they know I am? I can hear them in my ears, a scream, or a cry. A sigh of relief, as I look at the mirror and see them in front of me, like a mirror. You're gon na see me? ''                                                                                                                                                                                                                                                                                                                                                                                                                                                                                                                                                                                                                                                                                                                                                                                                                                                                                                                                                                                                                                                                                                                                                                                                                                                                                                                                                                                                                                                                                                                                                                                                                                                                                                                                                                                                                                                                                                     \\ \cmidrule{1-2}
		\textbf{BART\textsubscript{\textsc{large}}}       & I've always been a bit of a weirdo. \textless{}newline\textgreater I'm not sure why, but I always knew I was going to die at some point in my life. I don't know if it's because I was born in the wrong year, or if it was because my parents died when I was very young. I just knew that I would die. I've never been able to explain it to anyone, but it was always there in the back of my mind. It's not like I could do anything about it. I couldn't even do anything to stop it. It was just always there. I never knew what it was, or what it would do to me. I didn't want to know. I wanted it to be over. But it never was. It just kept getting worse and worse. I can't explain it, I just know that I 'll die.                                                                                                                                                                                                                                                                                                                                                                                                                                                                                                                                                                                                                                                                                                                                                                                                                                                                                                                                                                                                                                                                                                                                                                                                                                                                                                                                                                                                                                                                                                                                                                                                                                                                                                                                                                                                                                                                                                                \\ \cmidrule{1-2}
		\textbf{ProphetNet\textsubscript{\textsc{large}}} & it's been a long time since i've had a precognition, but i don't think i'll ever be able to get rid of it. \textless newline \textgreater it started when i was a kid. i was told that i had a very shitty type of pre - lucid dreaming. i had no idea what it was until i was in my late teens. it was like i could see the future in front of me. i knew exactly what i was going to do when i saw the future. i could feel it in my bones. i can feel it. i know exactly what's going to happen when i look at the future, and i know it will be the best thing that's ever happened to me. the future is going to be better than i ever thought it would be. i'm not going to let it happen again.                                                                                                                                                                                                                                                                                                                                                                                                                                                                                                                                                                                                                                                                                                                                                                                                                                                                                                                                                                                                                                                                                                                                                                                                                                                                                                                                                                                                                                                                                                                                                                                                                                                                                                                                                                                                                                                                                                                                          \\ \cmidrule{1-2}
		\textbf{Gold}       & Just looking at the guy I already knew it . He was in for a seven on the Bristol scale and it was only hours away . \textless{}newline\textgreater \textless{}newline\textgreater Looking at his carefree face from below as he was slowly transported towards me on the escalator , I wondered , not for the first time , if I should warn him . But as always I asked myself : How could I tell him ? How could I tell anyone ? \textless{}newline\textgreater \textless{}newline\textgreater Tell people that I could see the future . That I could see their fate ? That I could feel the forces that were at work , often without their knowledge and beyond their control , inside their own bodies ? That I could see shape of things to come ? That I was a coproscognitive ? \textless{}newline\textgreater \textless{}newline\textgreater Nobody would believe me . Not until after a few predictions had been proven to be correct at least and few humored me for long enough to realize the truth of my predictions . \textless{}newline\textgreater \textless{}newline\textgreater And this guy did n't have a few predictions to confirm the authenticy of my gift . He needed to know and believe me about the very next event I could foresee and there was no way to convince him in time . \textless{}newline\textgreater \textless{}newline\textgreater And thus I stayed silent again as the escalator carried us past one another leaving him to his fate . I was going up and he was going down . \textless{}newline\textgreater \textless{}newline\textgreater I could only hope for him that he would be close to the right place two hours and thirteen minutes from now , because I could tell it would be a fast and quite explosive one and without my word he would not have much advanced warning telling him to run for salvation . \textless{}newline\textgreater \textless{}newline\textgreater I put him out of my mind I had my own future to worry about and I did n't need my gift to tell me about it . I started looking for the signs pointing the way to closest lavatory .                        \\                                                                                                                                                                                                                                                                                                                                                                                                                                                                                                               
		\bottomrule                                                                                                                          
	\end{tabular}
	\caption{Test set examples of WritingPrompts dataset for each model.}
	\label{tab:writingprompts-ex1}
\end{table*}

\begin{table*}[]
	\small
	\centering
	\begin{tabular}{lp{0.8\textwidth}}
		\toprule
		\textbf{Prompts}    & {[} WP {]} `` Before we allow humankind to live among us in the stars , tell us the faults of your kind . ''                                                                                                                                                                                                                                                                                                                                                                                                                                                                                                                                                                                                                                                                                                                                                                                                                                                                                                                                                                                                                                                                                                                                                                                                                                                                                                                                                                                                                                                                                                                                                                                                                                                                                                                                                                                                                                                                                                                                                                                                                                                                                                                                                                                                                                                                                                                                                                                                                                                                                                                                    \\
		\midrule
		\textbf{GPT-2\textsubscript{\textsc{medium}}}       & \textless{}newline\textgreater `` I'm sorry, sir, but I don't think we're ready for this. '' I said. `` We've been here for thousands of years, and we've never seen anything like this before. '' He looked at me with a look of horror on his face. `` What do you mean? '' I asked. He looked down at his hands, and then back up at me. `` It's like this, '' he said, `` You can't see it, but it's there. It 'll be here for a long time, and it 'll kill us all. '' \textless newline\textgreater I looked at him, and he looked back at me, and both of us looked at the stars.                                                                                                                                                                                                                                                                                                                                                                                                                                                                                                                                                                                                                                                                                                                                                                                                                                                                                                                                                                                                                                                                                                                                                                                                                                                                                                                                                                                                                                                                                                                                                                                                                                                                                                                                                                                                                                                                                                                                                                                                                                                         \\
		\cmidrule{1-2}
		\textbf{UniLM\textsubscript{\textsc{large}}}      & ` ` we allow humanskind to be among us, tell them the faults.'\textless newline \textgreater \textless new new \textgreater ` ` humanskind,'' i said, ` ` you are the one who has been chosen to be our leader.'` ` humankind, you are our leader,'` i said. \textless new york. \textless york \textgreater \textless yoline \textgreater `'humanskind.'* \textless newrk \textgreater * \textless yo \textgreater \textless \textless new \textgreater \textless \textgreater humanskind : \textless new name \textgreater \textless nowline \textgreater humans : \textless now name \textgreater humans. \textless now names \textgreater \textless the name of the universe. \textless the names of the stars. \textless \textless{}line \textgreater * humanskind * \textless now named \textgreater \textless first name \textgreater the universe, \textless new names \textgreater the stars and the stars \textless new stars. the names \textless new planets. \textless first names \textgreater * * humans * \textless firstline \textgreater the name \textless new species \textgreater \textless humanline \textgreater \textgreater \textless humans \textgreater \textless last name \textgreater * humankind *. \textless last names \textgreater humans * * \textless lastline \textgreater humankind. \textless name \textgreater \textgreater * humanity * \textless name \textless \textgreater \textless * humans. * \textless * human * \textless \textless \textgreater \textgreater humans, \textless now known as humans. the name. \textless{}.. \textless * humanity. \textless human name \textgreater{}... * * * humanity \textless new humans \textgreater *.. humans *. *. humans.. '. \textless{}line. \textless humans.'s.. human.'the humans. they were the humans, the humans of the galaxy. \textless : \textless \textless \textless humans : humans. humans. humans, humans. humankind \textless new galaxy \textgreater \textless : / / www. reddit. com / r / writingprompts / comments / 2jclq / comments \_ 2jflq \_ comments / 1xxxfxxgxxcxxbxxdxxkxxqxx \textless new \_ \textgreater \textless \_ \_ \textless new i am humankind : the humankind of the galactic system. \textless \_ \textless \textless \_ \textgreater i am the human race. \textless tv \textgreater \textless tline \textgreater i was the human. \textless{}pline \textgreater it was a long time since i was human. i was a human.. i am a human race,..kind \textless{}. \textless{}'humans human \textless \textless race humans \textless * \textless human of * \textless the \textless \textless * i humans. new.. \\
		\cmidrule{1-2}
		\textbf{T5\textsubscript{\textsc{large}}}         & Before we allow humankind to live amongst the stars, tell us the faults of your kind. '' newline\textgreater I don't know, I'm not a scientist, but I do have a degree in astronomy, and I do know a thing or two about science. I know that a lot of people think that science is a good thing, that it's a great thing. But, if you think about it, you're a fucking shithole. You've got a bunch of crazies, all of them. So, what do you think?? Do you know what? I mean, they 'll tell you. And, of course, we 'd like to know what you think of us.                                                                                                                                                                                                                                                                                                                                                                                                                                                                                                                                                                                                                                                                                                                                                                                                                                                                                                                                                                                                                                                                                                                                                                                                                                                                                                                                                                                                                                                                                                                                                                                                                                                                                                                                                                                                                                                                                                                                                                                                                                                                                      \\
		\cmidrule{1-2}
		\textbf{BART\textsubscript{\textsc{large}}}       & `` Before we allow humankind to live among us in the stars, tell us the faults of your kind. '' \textless{}newline\textgreater \textless{}newlines\textgreater `` Well, first of all, they aren't very smart. They don't know how to read. They're not very good at math. They haven't learned how to write yet. They are also very lazy. They spend most of their time staring at their screens. They can't even get up to go to the bathroom. They just sit there and stare at the screen. They also have a tendency to stare at their phones for hours at a time. I'm not sure why they do that, but I guess it's because they're bored. ''\\
		\cmidrule{1-2}
		\textbf{ProphetNet\textsubscript{\textsc{large}}} & ` ` before we allow humankind to live among us in the stars, tell us the faults of our kind.'' \textless newline \textgreater ` ` i'm sorry, sir, but we don't have the technology to do that. we're too afraid of the consequences of our actions, and we've spent too much time trying to find a way to stop them.'cause they're just too stupid to do anything about it. we have to do something about it, or we'll never be able to get out of here. we need to find some way to get them out of there, and if they do, then we'd have to go back to earth and start all over again. and if that's the case, then i'd like to thank you for your time, and i hope to see you again soon,''                                                                                                                                                                                                                                                                                                                                                                                                                                                                                                                                                                                                                                                                                                                                                                                                                                                                                                                                                                                                                                                                                                                                                                                                                                                                                                                                                                                                                                                                                                                                                                                                                                                                                                                                                                                                                                                                                                                                                  \\
		\cmidrule{1-2}
		\textbf{Gold}       & Tell us your faults ? Really ? This was the question - the shibboleth - that unlocked the cosmos ? \textless{}newline\textgreater \textless{}newline\textgreater The Masters could have picked a scientist to answer but they feared she might mask ignorance . They could have picked from our global leaders bit they feared that they would mask deceit . They could have picked a holy man but feared he would mask violence , oppression , hate , intolerance ... the list of disqualifying sins was almost too long to enumerate . \textless{}newline\textgreater \textless{}newline\textgreater So they picked Josh Thornton , a 45 year old MBA in human resources . \textless{}newline\textgreater \textless{}newline\textgreater `` Our greatest weakness ? Well , I think we work a little too hard and , as a race , we might be a bit of a perfectionist .\\
		\bottomrule                                                                                                                          
	\end{tabular}
	\caption{Test set examples of WritingPrompts dataset for each model.}
	\label{tab:writingprompts-ex2}
\end{table*}


\end{document}